\documentclass[11pt]{article}

\usepackage{amssymb,amsmath,amsthm,bbm}
\usepackage{verbatim,float,url,dsfont}
\usepackage{graphicx,subcaption,psfrag}
\usepackage{algorithm}
\usepackage[noend]{algorithmic}
\usepackage{mathtools,enumitem}
\usepackage{multirow}
\usepackage{ragged2e}
\usepackage{xr-hyper}
\usepackage{caption}
\usepackage{array}

\usepackage[utf8]{inputenc} 
\usepackage[T1]{fontenc}    
\usepackage{booktabs}       
\usepackage{nicefrac}         
\usepackage{microtype}      

\usepackage[dvipsnames]{xcolor} 
\definecolor{hunyuanblue}{HTML}{1E4A8F}
\usepackage{pifont} 
\usepackage{etoc} 
\usepackage{tikz} 
\usepackage{pgfplots}  
\pgfplotsset{compat=1.16}  

\ifdefined\TimesFont 
\usepackage{times} 
\fi

\ifdefined\ParSkip 
\usepackage{parskip} 
\fi

\newtheorem*{assumption*}{\assumptionnumber}
\providecommand{\assumptionnumber}{}


\makeatletter
\newcommand*\rel@kern[1]{\kern#1\dimexpr\macc@kerna}
\newcommand*\widebar[1]{%
  \begingroup
  \def\mathaccent##1##2{%
    \rel@kern{0.8}%
    \overline{\rel@kern{-0.8}\macc@nucleus\rel@kern{0.2}}%
    \rel@kern{-0.2}%
  }%
  \macc@depth\@ne
  \let\math@bgroup\@empty \let\math@egroup\macc@set@skewchar
  \mathsurround\z@ \frozen@everymath{\mathgroup\macc@group\relax}%
  \macc@set@skewchar\relax
  \let\mathaccentV\macc@nested@a
  \macc@nested@a\relax111{#1}%
  \endgroup
}
\makeatother

\def\Id{\mathrm{Id}}

\usepackage{hunyuan}

\usepackage[capitalize,noabbrev]{cleveref}
\usepackage{tabularx}
\usepackage[most]{tcolorbox}
\usepackage{wrapfig}
\usepackage{xspace}
\usepackage{fancyhdr}
\usepackage{xurl}

\makeatletter
\long\def\@makecaption#1#2{%
  \vskip 10pt
  \setbox\@tempboxa\hbox{#1: #2}%
  \ifdim \wd\@tempboxa >\hsize
    \noindent #1: #2\par   
  \else
    \hbox to\hsize{\hfil\box\@tempboxa\hfil}
  \fi}
\makeatother

\hypersetup{
  colorlinks=true,
  linkcolor=hunyuanblue,
  citecolor=hunyuanblue,
  urlcolor=hunyuanblue,
}

\makeatletter
\def\section{\@startsiction{section}{1}{\z@}{-0.24in}{0.10in}
             {\large\bf\raggedright\color{hunyuanblue}}}
\def\subsection{\@startsection{subsection}{2}{\z@}{-0.20in}{0.08in}
                {\normalsize\bf\raggedright\color{hunyuanblue}}}
\makeatother

\fancypagestyle{plain}{%
  \fancyhf{}%
  \fancyfoot[C]{\thepage}%
}

\fancypagestyle{firststyle}{%
  \fancyhf{}%
  \fancyhead[L]{\raisebox{-18.5mm}[0pt][0pt]{\includegraphics[height=12mm]{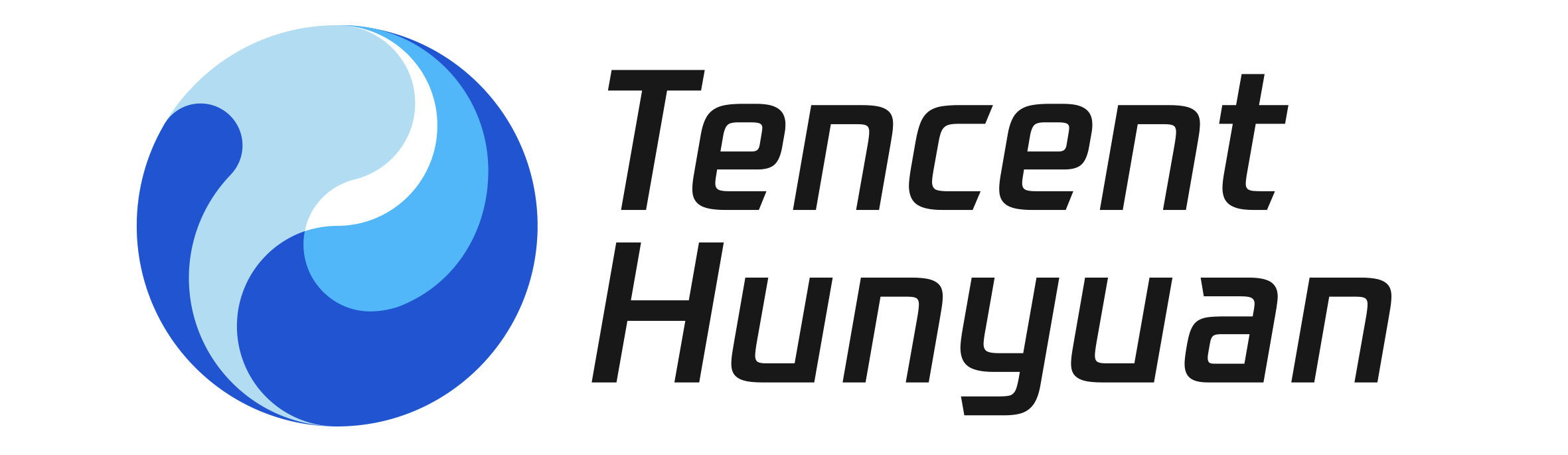}}}%
  \fancyfoot[C]{\thepage}%
}

\newcommand{\method}{DRPO\xspace}

\newcolumntype{Y}{>{\raggedright\arraybackslash}X}

\definecolor{caseblue}{RGB}{42, 91, 160}
\definecolor{casebg}{RGB}{246, 248, 252}
\definecolor{caseborder}{RGB}{180, 195, 220}
\definecolor{paperblue}{HTML}{1F77B4}
\definecolor{paperred}{HTML}{D62728}
\definecolor{deepred}{HTML}{B22222}
\definecolor{softred}{HTML}{C44E52}

\newtcolorbox[auto counter, number within=section]{compactcase}[2][]{
  breakable,
  enhanced,
  colback=gray!2,
  colframe=gray!30,
  colbacktitle=gray!12,
  coltitle=black, 1
  fonttitle=\bfseries,
  title={#2},
  boxrule=0.45pt,
  arc=1mm,
  left=1.5mm,
  right=1.5mm,
  top=1mm,
  bottom=1mm,
  toptitle=0.7mm,
  bottomtitle=0.7mm,
  before skip=0.8em,
  after skip=0.8em,
  label={#1}
}

\definecolor{abstractbg}{HTML}{F0F7FC}

\begin{document}

\thispagestyle{firststyle}
\vspace*{0.25cm}
{\color{hunyuanblue}\hrule height 0.6pt}
\vskip 6mm
\begin{center}
{\LARGE\bfseries Rethinking the Divergence Regularization in LLM RL\par}
\end{center}
\vskip 3mm
{\color{hunyuanblue}\hrule height 0.6pt}
\vskip 6mm
\begin{center}
\textbf{Jiarui Yao}$^{1,2,*}$ \quad
\textbf{Xiangxin Zhou}$^{1,*\,\mathparagraph}$ \quad
\textbf{Penghui Qi}$^{3,*\,\mathparagraph}$ \\[4pt]
\textbf{Wee Sun Lee}$^{3}$ \quad
\textbf{Liefeng Bo}$^{1}$ \quad
\textbf{Tianyu Pang}$^{1,\ddagger}$
\\[8pt]
$^1$Tencent Hunyuan \quad $^2$UIUC \quad $^3$NUS \\[6pt]
{\small $^*$Equal contribution \quad
$^\mathparagraph$Project Lead \quad
$^\ddagger$Corresponding author}
\end{center}
\vskip 6mm

\begin{tcolorbox}[
  colframe=abstractbg,
  colback=abstractbg,
  boxrule=0pt,
  arc=2mm,
  enhanced,
  top=12pt,
  bottom=12pt,
  left=15pt,
  right=15pt,
  width=\textwidth,
]
\textbf{Abstract.}\quad
Reinforcement learning (RL) has become a key component of post-training large language models (LLMs). In practice, LLM RL is often off-policy because of training-inference mismatch and policy staleness, making trust-region control essential for stable optimization. Mainstream methods such as PPO and GRPO approximate this control with a ratio-clipping mechanism, but the importance ratio can be a poor proxy for distributional shift in long-tailed vocabularies. Recent work such as DPPO addresses this mismatch by replacing ratio-based clipping with a divergence-based mask, yielding a trust region defined by the sampled token's absolute probability shift. However, DPPO still relies on a hard mask: once a token crosses the trust-region boundary in a harmful direction, its gradient is discarded rather than corrected. To address this, we propose \underline{D}ivergence \underline{R}egularized \underline{P}olicy \underline{O}ptimization (\method{}), which replaces the hard mask with a smooth advantage-weighted quadratic regularizer on policy shift. \method{} preserves the same trust-region geometry as DPPO while inducing bounded, continuous gradient weights that attenuate diverging updates and provide corrective signals beyond the boundary. Experiments across model scales, architectures, and precision settings show that \method{} improves the stability and efficiency of LLM RL training.

\vskip 8pt
\textbf{Date:} June 8, 2026 \\ 
\textbf{Code:} \url{https://github.com/Tencent-Hunyuan/UniRL/tree/main/DRPO}
\end{tcolorbox}


\section{Introduction}


Reinforcement learning (RL) has become a central component of post-training large language models (LLMs), enabling models to better align with human preferences and improve performance on complex reasoning tasks~\citep{ouyang2022training,rafailov2023direct,guo2025deepseekr1,drgrpo}. During training, an LLM is optimized as an autoregressive token-level policy that generates a response and receives a scalar reward from either a learned reward model~\citep{ouyang2022training} or a rule-based verifier~\citep{guo2025deepseekr1,yu2025dapo}.
In practice, modern LLM RL is typically \emph{off-policy}: rollouts are generated by inference engines whose numerical behavior differs from training engines \citep{fp16,yao2025offpolicy}, and collected trajectories are commonly split into multiple mini-batches or gradient steps~\citep{deepseek-v3.2}. As a result, the policy being updated is not identical to the behavior policy that generated the data.


In such off-policy settings, Trust Region Policy Optimization (TRPO) provides a principled solution by maximizing a surrogate objective under an explicit divergence constraint between the current and behavior policy~\citep{trpo,cpo}. However, its second-order optimization makes TRPO impractical to scale. Proximal Policy Optimization (PPO)~\citep{ppo} replaces the constrained optimization with a simple ratio-clipping heuristic and has become the dominant recipe in modern LLM RL training. Building on PPO, GRPO improves practicality by replacing a learned critic with group-relative reward normalization~\citep{grpo,rloo,drgrpo}. More recently, Simple Policy Optimization (SPO)~\citep{spo} replaces hard clipping with a smooth quadratic regularizer that preserves the same ratio boundary while avoiding the zero-gradient issue outside the clipping range. These methods differ in implementation details, but they share the same trust-region geometry: the per-token update is controlled through its importance ratio. \looseness=-1


The importance ratio, however, is a poor proxy for distributional shift for LLMs due to large and long-tailed vocabularies \citep{dppo}. A small increase on a low-probability token can produce a very large ratio while changing little probability mass. Conversely, a moderate ratio change on a high-probability token can move substantial mass and meaningfully alter the policy. A fixed ratio window therefore tends to over-constrain low-probability tokens and under-constrain high-probability tokens~\citep{dppo,yu2025dapo,cispo}. \looseness=-1

\begin{figure}[t]
    \centering
    \includegraphics[width=0.95\linewidth]{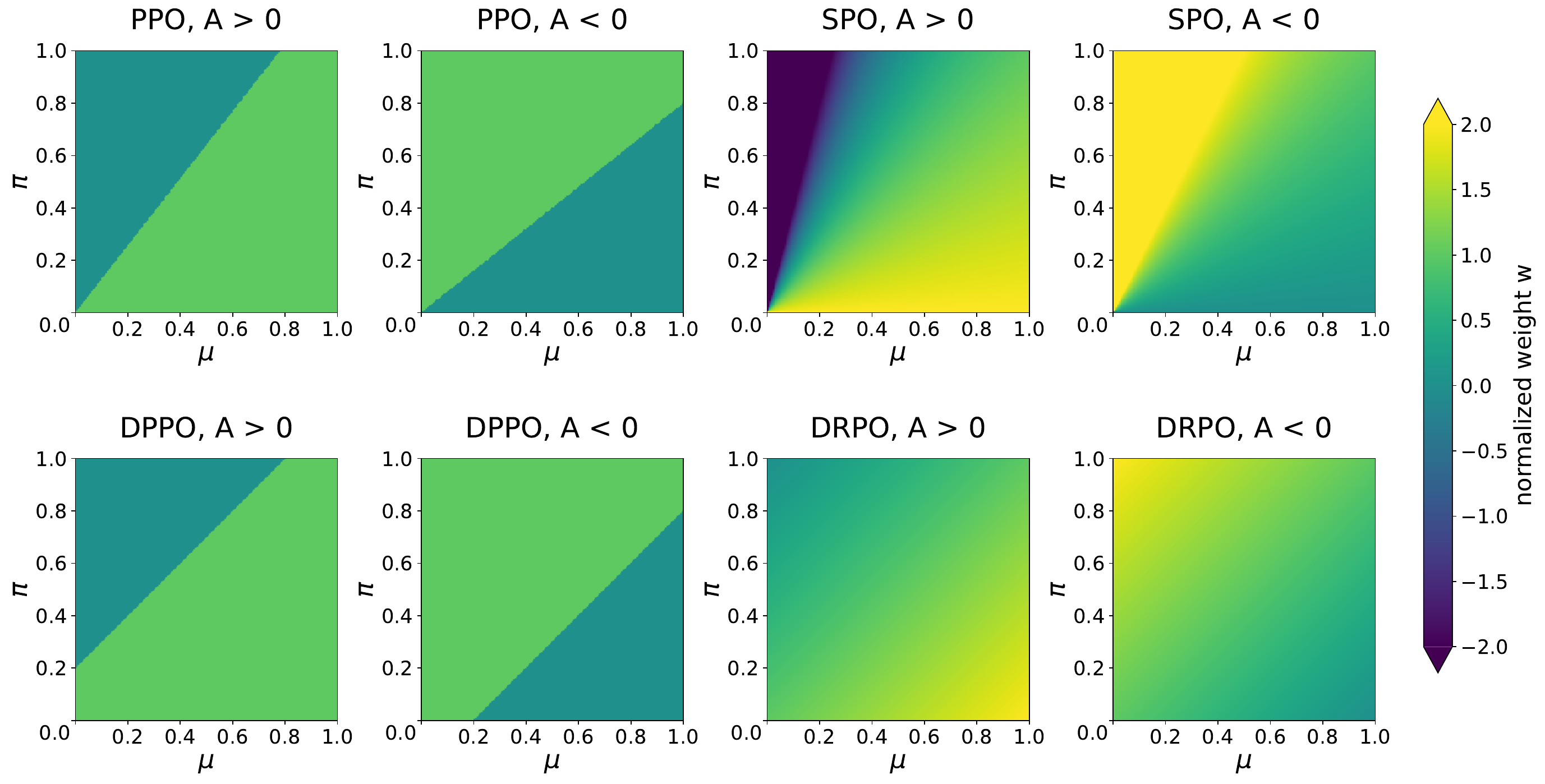}
    \caption{Per-token gradient weights of different algorithms as a function of the current probability $\pi(y_t|s_t)$ and behavior probability $\mu(y_t|s_t)$. For SPO, $\epsilon=1$; for \method{}, $\delta=1$; for PPO, $\varepsilon_{\rm low}=0.2$ and $\varepsilon_{\rm high}=0.28$; for DPPO, $\delta=0.2$. SPO's weight grows without bound as $\mu(y_t|s_t) \to 0$, while the weight of \method{} remains bounded for all tokens.}
    \label{fig:pg_weight_heatmaps}
\end{figure}


DPPO addresses this issue by replacing ratio-based clipping with a divergence-based mask~\citep{dppo}. When the policy divergence exceeds a prescribed threshold and the current update would increase it further, DPPO disables the corresponding token gradient.
Its Binary-TV variant, which we refer to as DPPO unless otherwise stated, measures the absolute probability shift of the sampled token. This quantity aligns more closely with total variation (TV) geometry than the importance ratio in long-tailed vocabularies. However, DPPO still enforces the trust region with a binary mask. Once a token moves outside the trust region in a harmful direction, its gradient is set to zero. This prevents further movement away from the behavior policy, but it provides no corrective signal to move the policy back toward the boundary and can introduce abrupt changes near the threshold.


We propose \method{}, a divergence-regularized policy optimization method that replaces the hard mask while preserving the Binary-TV trust region in DPPO. Our method is motivated by SPO, which places the per-token optimum exactly at PPO's trust-region boundary through an advantage-weighted $\chi^2$ regularizer. We rewrite the Binary-TV constraint as a token-adaptive ratio bound and apply the same construction as SPO, which yields an advantage-weighted $\ell_2^2$ regularizer. The resulting regularizer changes the trust-region geometry from a fixed ratio constraint to an absolute probability-shift constraint, combining the smoothness of SPO with the divergence-based geometry of DPPO.


Our method also gives a simple and stable gradient form. Each token's policy-gradient contribution is multiplied by a continuous weight determined by its Binary-TV shift and by whether the current update moves away from or toward the behavior policy. When the update moves away from the behavior policy, the weight decays to zero at the trust-region boundary and becomes corrective beyond it. When the update moves back toward the behavior policy, the weight is amplified. Because this weight depends on an absolute probability shift rather than an importance ratio, it better captures the geometry of policy change and remains bounded even in the low-probability tail where SPO's ratio-based weight can grow without bound.


Beyond the specific algorithm, our results motivate a gradient-centered view of regularizer design for LLM RL. Our ablations show that standard KL or TV penalties can underperform because their gradients reintroduce ratio-based geometry. They also show that the per-token penalty should be weighted by the absolute-advantage because it keeps the trust-region boundary independent of reward scale. These findings suggest three practical criteria for an effective regularizer: it should induce a stable boundary aligned with distributional shift, keep per-token gradient weights bounded in the long-tailed vocabulary, and provide a smooth corrective signal when the policy moves away. \method{} satisfies these criteria with a simple Binary-TV-aligned regularizer, offering an empirical lens for designing stable policy-optimization objectives for LLMs.

\section{Background}

The generation process of LLMs can be formulated as a token-level MDP \citep{bellman1957markovian} \(\mathcal{M} = (\mathcal{S}, \mathcal{A}, R, p_{\mathcal{X}})\). Given a prompt \(x \sim p_{\mathcal{X}}\), a response \(y=(y_1, \dots, y_T)\) is autoregressively sampled by a conditional stochastic policy \(\pi(y_t|s_t)\) over the vocabulary \(\mathcal{A}\), where the state \(s_t = (x, y_1, \dots, y_{t - 1}) \in \mathcal{S}\) is the concatenation of the prompt and the generated tokens so far. The generation terminates upon producing the \texttt{[eos]} token or reaching the token limit. A scalar reward \(R(x,y)\) is then provided, either from a reward model \citep{ouyang2022training} or a rule-based verifier \citep{guo2025deepseekr1}. The policy objective is to maximize the expected reward:
$$
\mathcal{J}(\pi) = \mathbb{E}_{x \sim p_{\mathcal{X}}}\left[ \mathcal{J}(x, \pi) \right] = \mathbb{E}_{x \sim p_{\mathcal{X}}}\left[ \mathbb{E}_{y \sim \pi(\cdot|x)}[ R(x,y) ] \right].
$$

Modern RL frameworks for LLM fine-tuning rely on highly optimized training and inference engines to maximize throughput, which inevitably introduces subtle but non-negligible numerical discrepancies \citep{fp16,yao2025offpolicy}. A further common practice is to collect a large batch of rollouts and split it into multiple mini-batches for multiple gradient updates \citep{deepseek-v3.2}. Both cases bring RL training into an \emph{off-policy} paradigm, where the data is sampled from a behavior policy $\mu$ and the objective becomes:
\begin{equation}\label{eq:seq_is_obj}
    \mathcal{J}(x, \pi) = \mathbb{E}_{y \sim \mu(\cdot|x)} \left[ \prod_{t=1}^{|y|} \frac{\pi(y_t|s_t)}{\mu(y_t|s_t)} \cdot R(x,y) \right].
\end{equation}

\subsection{Trust Region Policy Optimization}

Directly optimizing \Cref{eq:seq_is_obj} often suffers from high variance due to the product of importance sampling ratios. TRPO \citep{trpo} handles this with a token-level surrogate objective:
\begin{equation}\label{eq:tok_is_obj}
    \mathcal{L}(x, \pi) = \mathbb{E}_{y \sim \mu(\cdot|x)} \left[ \sum_{t=1}^{|y|} \frac{\pi(y_t|s_t)}{\mu(y_t|s_t)} \cdot \hat{A}_t \right],
\end{equation}
where $\hat{A}_t = R(x,y) - V(s_t)$ is the advantage estimate, and $V(s_t)$ is a variance-reduction baseline that does not change the expected policy gradient. Typically, $V(s_t)$ is set to the expected reward conditioned on state $s_t$.
TRPO and later work \citep{cpo,dppo} have shown that this surrogate is a first-order approximation\footnote{We adapt TRPO to the LLM setting and ignore a constant term; see \citet{dppo} for a rigorous derivation.} of \Cref{eq:seq_is_obj}, and a monotonic performance improvement can be guaranteed within a \textit{trust region} defined by the KL divergence or TV distance. Formally, TRPO solves the following constrained optimization problem:
\begin{equation}\label{eq:trpo-obj}
\begin{split}
    \max_{\pi} \,\,  \mathcal{L}(x, \pi) \quad\quad
    \text{s.t.} \,\,  \mathbb{E}_{y \sim \mu(\cdot|x)}\left[ \sum_{t=1}^{|y|} D_{\mathrm{TV}} \left(\mu(\cdot|s_t) \| \pi(\cdot|s_t) \right)  \right] \le \delta.
\end{split}
\end{equation}

\subsection{Proximal Policy Optimization}

TRPO requires second-order methods that are computationally prohibitive at scale. PPO \citep{ppo} was introduced as a simple alternative that approximates the trust region via a ratio-clipping mechanism. Letting $r_t \triangleq \frac{\pi(y_t|s_t)}{\mu(y_t|s_t)}$ denote the per-token importance ratio, PPO optimizes:
\begin{equation}\label{eq:ppo_obj}
    \mathcal{L}_\mathrm{PPO}(x, \pi) = \mathbb{E}_{y \sim \mu(\cdot|x)} \left[ \sum_{t=1}^{|y|} \min \!\left( r_t \cdot \hat{A}_t,\; \operatorname{clip}( r_t, 1 - \epsilon, 1 + \epsilon) \cdot \hat{A}_t \right) \right].
\end{equation} 
The clipping mechanism deactivates the gradient whenever $r_t$ leaves the interval $[1-\epsilon,\,1+\epsilon]$ and further increase the loss, thereby enforcing a per-token, ratio-based trust region, i.e.,
\(
    |r_t - 1| \le \epsilon.
\)

\noindent\textbf{Group Relative Policy Optimization.} In traditional RL settings, $V(s_t)$ is typically estimated by a critic model. Learning such a critic is, however, expensive and noisy for LLMs. To address this, \citet{grpo,rloo,drgrpo} propose sampling a group of responses $\{y_i\}_{i=1}^G$ per prompt and estimating the advantage as $\hat{A}_{t,i} = R(x,y_i) - \frac{1}{G}\sum_{j=1}^{G} R(x,y_j)$. This critic-free approach is widely known as Group Relative Policy Optimization (GRPO).

\subsection{Simple Policy Optimization}

While effective in practice, PPO enforces its trust region through a hard clipping rule. This mechanism is brittle near the clipping boundary: a small change in $r_t$ can abruptly switch a token's gradient from active to zero. Moreover, once a token has moved outside the clip range in a harmful direction, PPO removes its gradient entirely and provides no corrective signal back toward the trust region. SPO \citep{spo} addresses these issues by replacing the hard clip with a smooth quadratic regularizer:
\begin{equation}\label{eq:spo_obj}
    \mathcal{L}_\mathrm{SPO}(x, \pi) = \mathbb{E}_{y \sim \mu(\cdot|x)} \left[ \sum_{t=1}^{|y|} \bigg( r_t \cdot \hat{A}_t - \frac{|\hat{A}_t|}{2\epsilon}\,(r_t - 1)^2 \bigg) \right].
\end{equation}
For each token, the integrand is a concave quadratic in $r_t$. Setting its derivative $\hat{A}_t - \frac{|\hat{A}_t|}{\epsilon}(r_t - 1)$ to zero gives the unique maximizer $r_t^\star = 1 + \operatorname{sign}(\hat{A}_t)\epsilon$, which exactly matches PPO's clipping boundary in \Cref{eq:ppo_obj}. SPO therefore preserves the same ratio-based trust region as PPO, but enforces it through a continuous gradient weight.

\subsection{Divergence Proximal Policy Optimization}

PPO, GRPO, and SPO all derive their trust region from the per-token ratio $r_t$. \citet{dppo} argues that this estimator is poorly behaved over LLMs' long-tailed vocabulary: a low-probability token can produce an enormous ratio (e.g., $10^{-5} \!\to\! 10^{-3}$) while contributing negligibly to the actual distributional shift, whereas a high-probability token may exhibit a modest ratio (e.g., $0.99 \!\to\! 0.80$) that nevertheless induces a substantial change in policy. Ratio-based trust regions thus over-penalize low-probability tokens, which are often exploratory, and under-penalize high-probability ones, harming both efficiency and stability.

DPPO \citep{dppo} replaces the ratio-based clip with a divergence-based mask $M_t^\mathrm{DPPO}$ conditioned on the policy divergence $D_t \triangleq D\big(\mu(\cdot|s_t)\,\|\,\pi(\cdot|s_t)\big)$, where $D$ is either the TV or KL divergence over the full per-state token distributions. The DPPO objective and mask are
\begin{equation}\label{eq:dppo_obj}
\begin{aligned}
    \mathcal{L}_\mathrm{DPPO}(x, \pi) &= \mathbb{E}_{y \sim \mu(\cdot|x)} \left[ \sum_{t=1}^{|y|} M_t^\mathrm{DPPO} \cdot r_t \cdot \hat{A}_t \right], \\
    M_t^\mathrm{DPPO} &=
    \begin{cases}
        0, & \operatorname{sign}\!\big(\hat{A}_t \cdot (r_t - 1)\big) > 0 \text{ and } D_t > \delta, \\
        1, & \text{otherwise},
    \end{cases}
\end{aligned}
\end{equation}
with divergence threshold $\delta$. The mask zeros the gradient only when the policy has already moved outside the trust region in a direction that would push it further away. For tractability over large vocabularies, DPPO approximates $D_t$ with binary or top-$k$ surrogates. Most relevant to our method is the Binary-TV approximation, which collapses the per-state distribution into a Bernoulli over the sampled token versus the rest, yielding
\begin{equation}\label{eq:divergence_bin_tv}
    D_t^\mathrm{Bin\text{-}TV} \,\triangleq\, \big|\pi(y_t|s_t) - \mu(y_t|s_t)\big|.
\end{equation}
The corresponding trust region $\big|\pi(y_t|s_t) - \mu(y_t|s_t)\big| \le \delta$ constrains the \emph{absolute} probability shift on the sampled token, in contrast to the \emph{relative} ratio constraint $|r_t - 1| \le \epsilon$ shared by PPO and SPO. 

\section{Method}\label{sec:method}

We derive \method{} from the Binary-TV view of DPPO. For a sampled token $y_t$, the Binary-TV proxy in \Cref{eq:divergence_bin_tv} satisfies
\(
    D_t^{\mathrm{Bin\text{-}TV}}
    = \big|\pi(y_t|s_t)-\mu(y_t|s_t)\big|
    = \mu(y_t|s_t)\,|r_t-1|.
\)
Thus the Binary-TV trust region $D_t^{\mathrm{Bin\text{-}TV}} \le \delta$ is equivalent to a token-adaptive ratio constraint,
\(
    |r_t - 1| \le \frac{\delta}{\mu(y_t|s_t)}.
\)
Under this view, DPPO can be represented by a PPO-style clipped surrogate with the same gradient behavior:
\begin{equation*}
    \mathcal{L}_\mathrm{DPPO}(x, \pi) =
    \mathbb{E}_{y \sim \mu(\cdot|x)} \!\left[ \sum_{t=1}^{|y|} \min \!\left( r_t \hat{A}_t,\; \operatorname{clip}\left( r_t, 1 - \frac{\delta}{\mu(y_t|s_t)}, 1 + \frac{\delta}{\mu(y_t|s_t)}\right) \hat{A}_t \right) \right].
\end{equation*}
Compared with PPO in \Cref{eq:ppo_obj}, DPPO replaces the fixed ratio interval with an adaptive one whose width is inversely proportional to the behavior probability of the sampled token. Low-probability tokens therefore receive a looser ratio tolerance, while high-probability tokens receive a tighter one. This constraint avoids the main failure mode of ratio-based trust regions, which can over-penalize rare tokens and under-penalize common ones \citep{dppo}.

However, DPPO still enforces this divergence-based trust region through a binary mask, which makes the update brittle near the boundary: a small change in the estimated divergence can abruptly switch a token's gradient from full strength to zero. The key lesson from SPO is that the same trust-region boundary can be enforced by a smooth regularizer instead of a discontinuous cutoff. Such a regularizer induces a continuous gradient weight that varies with both the magnitude and direction of the probability shift. Inside the boundary, it smoothly reweights the policy gradient; outside the boundary, it provides a corrective mechanism that can pull the policy back toward the trust region. We apply this principle to the Binary-TV trust region by replacing DPPO's mask with a quadratic regularizer on the sampled token's absolute probability shift. The resulting objective, \emph{Divergence Regularized Policy Optimization} (\method{}), is
\begin{equation}\label{eq:our_objective_mc}
    \mathcal{L}_\text{\method}(x,\pi) =
    \mathbb{E}_{y \sim \mu(\cdot|x)}\!\left[\sum_{t=1}^{|y|}  r_t \hat{A}_t - \frac{|\hat{A}_t|}{2 \delta}\,\textcolor{red}{\mu(y_t|s_t)}\,(r_t - 1)^2 \right].
\end{equation}
The first term is the token-level surrogate in \Cref{eq:tok_is_obj}. The second term is a quadratic regularizer whose curvature is scaled by the behavior probability of the sampled token. This single factor changes the equilibrium from a fixed ratio shift, as in PPO and SPO, to a fixed absolute probability shift, as required by DPPO.
Taking the gradient of \Cref{eq:our_objective_mc} gives (see Appendix \ref{app:full_grad_derivation} for a full derivation)
\begin{equation}\label{eq:our_grad}
\begin{split}
    \nabla \mathcal{L}_\text{\method}(x,\pi) 
    ={} \mathbb{E}_{y \sim \mu(\cdot|x)}\!\left[\sum_{t=1}^{|y|}
        \left(1 - \operatorname{sign}(\hat{A}_t(r_t - 1))\frac{D_t^{\mathrm{Bin\text{-}TV}}}{\delta}\right)
        r_t \hat{A}_t \nabla \log \pi(y_t|s_t)
    \right].
\end{split}
\end{equation}
Relative to the unregularized gradient of \Cref{eq:tok_is_obj}, \method{} multiplies each token's policy-gradient contribution by a continuous weight
\begin{equation}\label{eq:adaptive_weight}
    w_t
    = 1 - \operatorname{sign}(\hat{A}_t(r_t - 1))\,\frac{D_t^{\mathrm{Bin\text{-}TV}}}{\delta}.
\end{equation}
The sign term indicates whether the current update moves the sampled probability away from or toward the behavior policy. The magnitude term measures the Binary-TV shift that should be controlled. Together, these terms make the weight vary smoothly with both the size and direction of the sampled token's probability shift.

\begin{table}[h]
    \centering
    \caption{Comparison of trust-region mechanisms. DPPO and \method{} enforce a Binary-TV constraint on the sampled token's absolute probability shift. Because TV is bounded in $[0,1]$, \method{} produces bounded gradient weights, whereas the ratio-based constraint in SPO does not.}
    \label{tab:method_comparison}
    \resizebox{\linewidth}{!}{%
    \begin{tabular}{llllc}
    \toprule
    Method & Mechanism & Trust-region constraint & Gradient weight $w_t$ & Range of $w_t$ \\
    \midrule
    PPO       & hard clip      & $|r_t - 1| \le \epsilon$              & $0$ or $1$ & $\{0,1\}$ \\
    SPO       & smooth regularizer & $|r_t - 1| \le \epsilon$              & $1 - \operatorname{sign}(\hat{A}_t(r_t{-}1))\,|r_t-1|/\epsilon$ & $(-\infty,+\infty)$ \\
    DPPO   & hard mask      & $|\pi(y_t|s_t) - \mu(y_t|s_t)| \le \delta$ & $0$ or $1$ & $\{0,1\}$ \\
    \method{} & smooth regularizer & $|\pi(y_t|s_t) - \mu(y_t|s_t)| \le \delta$ & $1 - \operatorname{sign}(\hat{A}_t(r_t{-}1))\,|\pi(y_t|s_t)-\mu(y_t|s_t)|/\delta$ & $[1-\tfrac{1}{\delta},\,1+\tfrac{1}{\delta}]$ \\
    \bottomrule
    \end{tabular}}
\end{table}

\subsection{Trust Region Analysis}\label{sec:trust_region_analysis}

We now examine how the smooth gradient weight in \Cref{eq:adaptive_weight} encodes the trust-region boundary.

\noindent\textbf{Diverging update ($\operatorname{sign}(\hat{A}_t(r_t - 1)) > 0$).}
When the update moves $\pi(y_t|s_t)$ away from $\mu(y_t|s_t)$, the weight becomes
\(
    w_t = 1 - D_t^{\mathrm{Bin\text{-}TV}}/\delta.
\)
Thus the gradient is gradually attenuated as the Binary-TV shift approaches the boundary. Inside the trust region, where $D_t^{\mathrm{Bin\text{-}TV}} < \delta$, the weight remains positive and the update still follows the reward-improving direction. Outside the trust region, where $D_t^{\mathrm{Bin\text{-}TV}} > \delta$, the weight is negative, so the gradient reverses and provides a corrective signal back toward the trust region. Since the per-token objective in \Cref{eq:our_objective_mc} is a concave quadratic in $r_t$, the zero-weight condition gives the stationary point
\begin{equation}\label{eq:adaspo_optim}
    \pi(y_t|s_t)^\star = \mu(y_t|s_t) + \operatorname{sign}(\hat{A}_t)\,\delta,
\end{equation}
which matches DPPO's trust region boundary when the same threshold $\delta$ is used.

\noindent\textbf{Converging update ($\operatorname{sign}(\hat{A}_t(r_t - 1)) < 0$).}
When the update moves $\pi(y_t|s_t)$ toward $\mu(y_t|s_t)$, the weight becomes
\(
    w_t = 1 + D_t^{\mathrm{Bin\text{-}TV}}/\delta.
\)
The gradient is therefore amplified rather than suppressed, encouraging the policy to move smoothly back toward the behavior policy.

\noindent\textbf{Takeaway.}
The two cases show that \textbf{\method{} preserves the same trust-region boundary as DPPO when the same threshold $\delta$ is used}, but replaces the brittle hard mask with continuous gradient reweighting. Inside the boundary, tokens continue moving in the reward-improving direction with smoothly attenuated gradients. Outside the boundary, the gradient reverses and provides a corrective signal back toward the trust region.

\subsection{Comparison with SPO}\label{sec:divergence_comparison}

To justify why the probability factor in \Cref{eq:our_objective_mc} is essential, we compare \method{} and SPO from two perspectives: the divergence each method implicitly regularizes, and the stability of the resulting per-token gradient weight. \Cref{tab:method_comparison} summarizes the key design differences across the four objectives.

\noindent\textbf{Implicit regularizer: $\ell_2^2$ versus $\chi^2$.}
For a fixed state $s_t$, write $\hat A_t(a)$ for the advantage that would be assigned when the sampled token is $a$. The regularization term in \method{} has expectation
\begin{equation*}\label{eq:l2_mc}
    \mathbb{E}_{y_t \sim \mu(\cdot|s_t)}\!\left[
        |\hat{A}_t(y_t)|\,\mu(y_t|s_t)(r_t-1)^2
    \right]
    =
    \sum_{a \in \mathcal{A}}
        |\hat{A}_t(a)|\,\bigl(\pi(a|s_t)-\mu(a|s_t)\bigr)^2 .
\end{equation*}
Thus \method{} penalizes an advantage-weighted squared $\ell_2$ distance between $\pi(\cdot|s_t)$ and $\mu(\cdot|s_t)$. In contrast, SPO uses the same quadratic form without the factor $\mu(y_t|s_t)$, giving
\begin{equation*}\label{eq:chi2_mc}
    \mathbb{E}_{y_t \sim \mu(\cdot|s_t)}\!\left[
        |\hat{A}_t(y_t)|\,(r_t-1)^2
    \right]
    =
    \sum_{a \in \mathcal{A}}
        |\hat{A}_t(a)|\,
        \frac{\bigl(\pi(a|s_t)-\mu(a|s_t)\bigr)^2}{\textcolor{red}{\mu(a|s_t)}} .
\end{equation*}
This is an advantage-weighted Pearson-$\chi^2$ penalty. The advantage weights modulate which tokens matter more for learning, but the key geometric difference comes from the denominator $\mu(a|s_t)$. SPO scales each squared probability shift by $1/\mu(a|s_t)$, making the penalty highly sensitive to deviations on low-probability tokens. \method{} instead penalizes the absolute probability shift directly: at a fixed advantage value, the same shift $|\pi(a|s_t)-\mu(a|s_t)|$ receives the same cost regardless of the token's behavior probability. In this sense, the $\ell_2^2$-type penalty is symmetric in $\pi$ and $\mu$, whereas the $\chi^2$-type penalty is tied to the behavior policy and can be dominated by the low-probability tail of $\mu$.

\begin{figure}[h]
    \centering
    \includegraphics[width=0.95\linewidth]{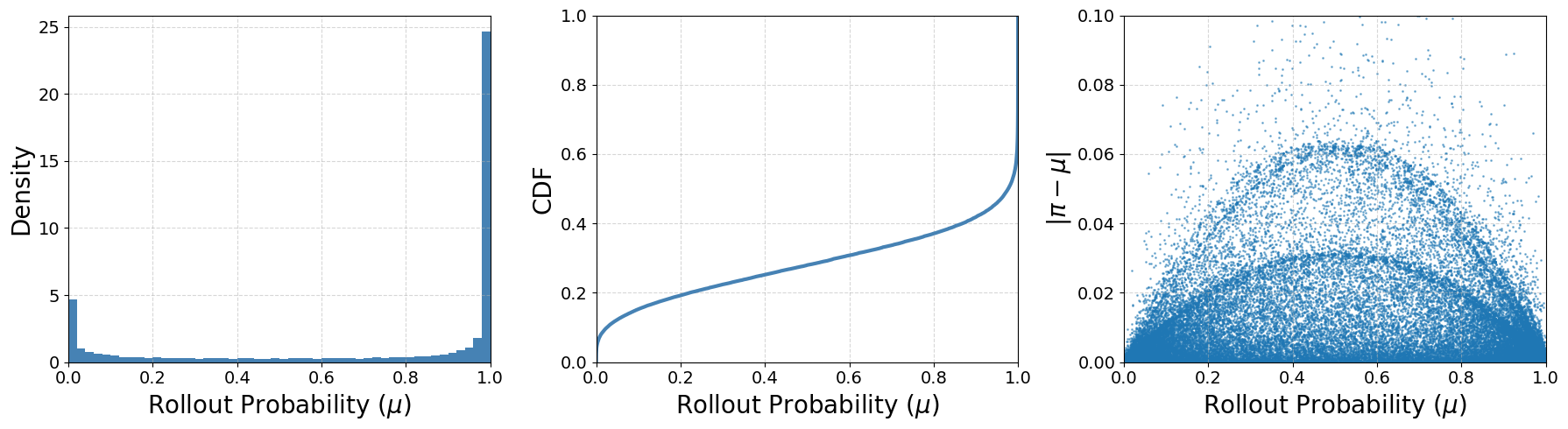}
    \caption{Histogram, cumulative distribution, and absolute probability shift $|\pi-\mu|$ of rollout probabilities $\mu(y_t|s_t)$ for tokens sampled from Qwen3-30B-A3B-Base \citep{qwen3}. The shift $|\pi-\mu|$ reflects training-inference mismatch. Tokens with $\mu(y_t|s_t) \le 0.01$ account for 7.8\% of all sampled tokens, showing that the low-probability tail is sampled non-negligibly often.}
    \label{fig:rollout_prob_hist_cdf}
\end{figure}

\noindent\textbf{Gradient stability in the long tail.}
A similar distinction appears in the gradient weights. From \Cref{tab:method_comparison}, SPO weights each token by a term involving
\(
    |r_t - 1|.
\)
Under $y_t \sim \mu(\cdot|s_t)$, this quantity is an unbiased single-sample Monte Carlo estimator of the unnormalized TV distance \citep{dppo}:
\begin{equation*}\label{eq:rt_tv_identity}
    \mathbb{E}_{y_t \sim \mu(\cdot|s_t)}\!\left[|r_t-1|\right]
    = \sum_{a \in \mathcal{A}} |\pi(a|s_t) - \mu(a|s_t)|
    = 2\,D_{\mathrm{TV}}\!\bigl(\mu(\cdot|s_t)\,\|\,\pi(\cdot|s_t)\bigr).
\end{equation*}
Its variance, however, is
\[
    \operatorname{Var}_{y_t \sim \mu(\cdot|s_t)}\!\left(|r_t-1|\right)
    =
    \chi^2\!\left(\pi(\cdot|s_t)\,\|\,\mu(\cdot|s_t)\right)
    -
    \left(2D_{\mathrm{TV}}\!\bigl(\mu(\cdot|s_t)\,\|\,\pi(\cdot|s_t)\bigr)\right)^2 .
\]
The $\chi^2$ term contains the factor $1/\mu(a|s_t)$, so the variance can become arbitrarily large when probability mass shifts on tokens with very small behavior probability. This is the typical long-tail regime of LLM sampling. As \Cref{fig:rollout_prob_hist_cdf} shows, tokens with $\mu(y_t|s_t)\le 0.01$ account for 7.8\% of all sampled tokens. For these tokens, even a modest absolute probability shift can induce a large ratio change, causing the SPO weight $1\pm |r_t-1|/\epsilon$ to dominate the gradient despite a small contribution to the actual distributional shift.

\method{} avoids this instability by replacing $|r_t-1|$ with
\(
    |\pi(y_t|s_t)-\mu(y_t|s_t)|,
\)
which directly measures absolute probability shift and more faithfully reflects the geometry of TV divergence (compare the right panel of \Cref{fig:rollout_prob_hist_cdf} with Figure~1 of \citet{dppo}). Since it is bounded in $[0,1]$ for every token, its variance is bounded by $1/4$, and the gradient weight of \method{} is confined to
\(
    1-\frac{1}{\delta} \le w_t \le 1+\frac{1}{\delta}.
\)
\Cref{fig:pg_weight_heatmaps} illustrates this contrast. SPO's weight grows without bound along the low-$\mu$ axis, whereas \method{} remains bounded everywhere. Thus \method{} realizes a smooth version of DPPO's divergence-based trust region while avoiding the high-variance weighting induced by ratio-based regularization.

\section{Experiments and Results}
\label{sec:experiment}

\noindent\textbf{Models, Data, and Benchmarks.}
We perform RL fine-tuning on Qwen3-4B-Base, Qwen3-30B-A3B-Base, and Qwen3.5-35B-A3B-Base~\citep{qwen3}, using a filtered subset of the original DAPO dataset~\citep{yu2025dapo} that contains approximately 13K math problems with rule-based verification. In addition, we fine-tune DeepSeek-R1-Distill-Qwen-1.5B (R1D)~\citep{guo2025deepseekr1} on a small sanity test dataset of 1,460 solvable questions~\citep{fp16}. During training, we evaluate on AIME 2024 and AIME 2025~\citep{AIME}. For each problem, we sample 16 responses and report the average score.

\noindent\textbf{Experimental Settings.}
We use the VeRL framework~\citep{sheng2024hybridflow} for RL training, with BF16 precision by default. For Qwen3-30B-A3B-Base, we additionally consider two low-precision settings: FP8 for rollout only, and FP8 for both training and rollout (FP8-E2E). These settings make optimization more challenging because FP8 precision, together with the MoE architecture, can increase the numerical mismatch between training and inference. Across all settings, we evaluate the unregularized trust-region-free surrogate (\Cref{eq:tok_is_obj}), GRPO~(\Cref{eq:ppo_obj}), SPO~(\Cref{eq:spo_obj}), DPPO~(\Cref{eq:dppo_obj}), and our proposed \method{}~(\Cref{eq:our_objective_mc}). For GRPO, we adopt the clip-higher trick with $\epsilon_\text{low}=0.2$ and $\epsilon_\text{high}=0.28$, following \citet{yu2025dapo}. For DPPO, we use the recommended value $\delta=0.15$. For SPO and \method{}, we set the regularization threshold to 12.5.
For other hyperparameters and hardware requirements, please refer to Appendix~\ref{sec:app:experiment} and Table~\ref{tab:hyperparameters}.

\begin{figure}[h]
    \centering
    \includegraphics[width=\linewidth]{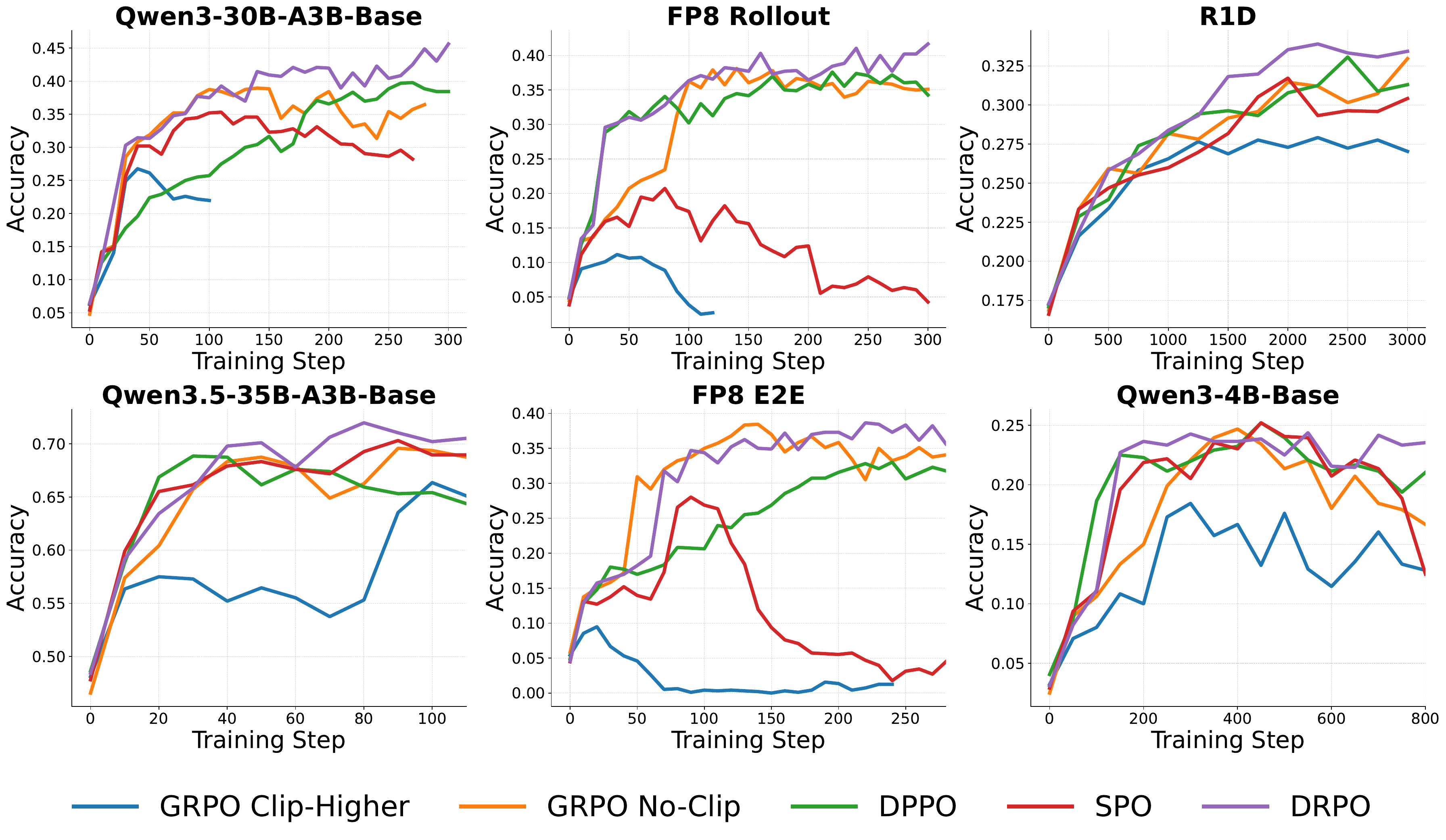}
    \caption{Average accuracy across all main experiment settings on AIME24 and AIME25.}
    \label{fig:main_accuracy}
\end{figure}

\subsection{Main Results}\label{sec:main_results}

We present the main results in \Cref{fig:main_accuracy} (see Appendix~\ref{app:comparing_kl} for comparing with KL regularization). Across all six settings, our \method{} consistently enables stable and efficient training, matching or exceeding the best evaluation accuracy achieved by the baselines.

\noindent\textbf{Instability of ratio-based methods.}
We find that ratio-based methods, namely GRPO and SPO, generally suffer from unstable training. This issue is especially severe in the low-precision settings, where they often collapse before reaching reasonable performance. Even in their strongest settings, their training efficiency and final accuracy lag behind their divergence-based counterparts. This observation is consistent with \citet{dppo}, which shows that $|r_t - 1|$ is a poor proxy for the true divergence and that ratio-based trust regions can lead to unstable and inefficient optimization.

\noindent\textbf{Limitations of a hard mask.}
Another observation is that hard-mask methods, such as GRPO and DPPO, often underperform their counterparts with smooth regularization. For example, although DPPO trains stably on Qwen3-30B-A3B-Base, it often converges more slowly and reaches lower final accuracy than \method{}. This supports our main claim that a smooth gradient signal is more effective in practice than a brittle hard mask.

\noindent\textbf{The need for a proper trust region.}
In some cases, the unregularized trust-region-free surrogate in \Cref{eq:tok_is_obj} already achieves strong performance, while a hard mask or ratio-based trust region can degrade performance. However, this unregularized surrogate is not reliable across settings, suffering a performance drop in three of the six settings. The most notable example is in the Qwen3-4B-Base experiment, where the accuracy decreases from 0.25 to 0.17. These results support the claim of \citet{dppo} that a trust region remains necessary, but suggest that its form is crucial.

Overall, \method{} combines the stability of divergence-based trust regions with the flexibility of a smooth regularizer, yielding the best overall performance across our experiments.

\subsection{Ablation Studies}\label{sec:ablation_study}

\begin{figure}[h]
    \centering
    \includegraphics[width=0.8\linewidth]{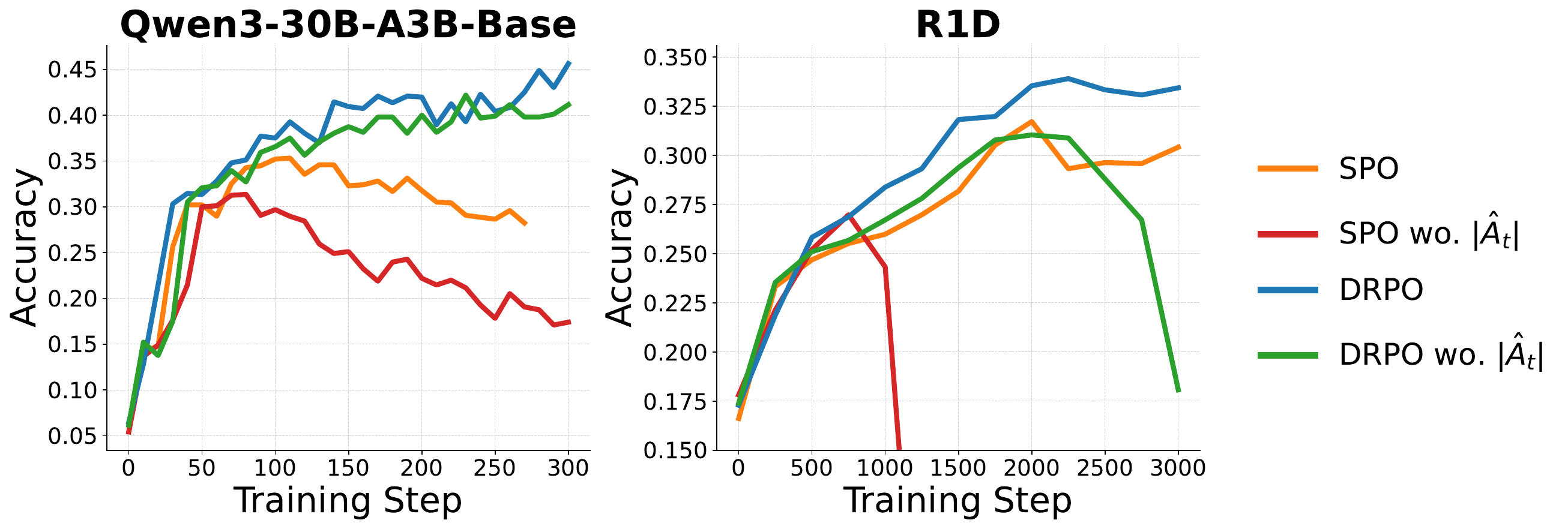}
    \caption{Ablation on $|\hat{A}_t|$. Removing this term degrades performance and destabilizes training.}
    \label{fig:ablation_adv}
\end{figure}

To further evaluate the effectiveness of our proposed method, we conduct a series of ablation studies on the design considerations of the regularizer.

\noindent\textbf{Advantage weight.}
In both SPO and \method{}, the regularization term is weighted by the absolute advantage $|\hat{A}_t|$. This weighting ensures that the per-token optimum lies on a stable trust-region boundary that does not depend on the magnitude of the advantage. Without this weighting, the trust-region boundary in \Cref{eq:adaspo_optim} would be coupled with $|\hat{A}_t|$, making it sensitive to token-level advantage noise and group-level advantage variance. However, this choice also makes the regularizer advantage-weighted rather than a pure divergence, as used in many prior works~\citep{r2vpo,becker2025troll}.

To examine whether $|\hat{A}_t|$ is necessary, we conduct ablations on Qwen3-30B-A3B-Base FP8-E2E and R1D by removing this factor from SPO and \method{} (see Appendix~\ref{app:extended_adv_weight} for this ablation on other alternative regularizations). As shown in \Cref{fig:ablation_adv}, removing $|\hat{A}_t|$ consistently causes a performance drop and leads to training instability. These results suggest that maintaining a stable trust-region boundary is more important than enforcing a pure divergence form for the regularizer.
This behavior is reasonable because $|\hat{A}_t|$ also determines the scale of the per-token policy gradient. Scaling the regularizer by $|\hat{A}_t|$ preserves the same relative corrective strength across tokens with different advantage magnitudes. Without this scaling, tokens with small advantages can be over-regularized, while tokens with large advantages can move too far before receiving sufficient correction. \looseness=-1

\noindent\textbf{Other alternative regularizations.}
As shown in \Cref{sec:divergence_comparison}, the regularizer in \method{} can be interpreted as an advantage-weighted $\ell_2^2$ penalty, whereas the regularizer in SPO corresponds to an advantage-weighted $\chi^2$ divergence. This raises a natural question: can other divergence measures yield better performance?

\begin{figure}[h]
    \centering
    \includegraphics[width=0.8\linewidth]{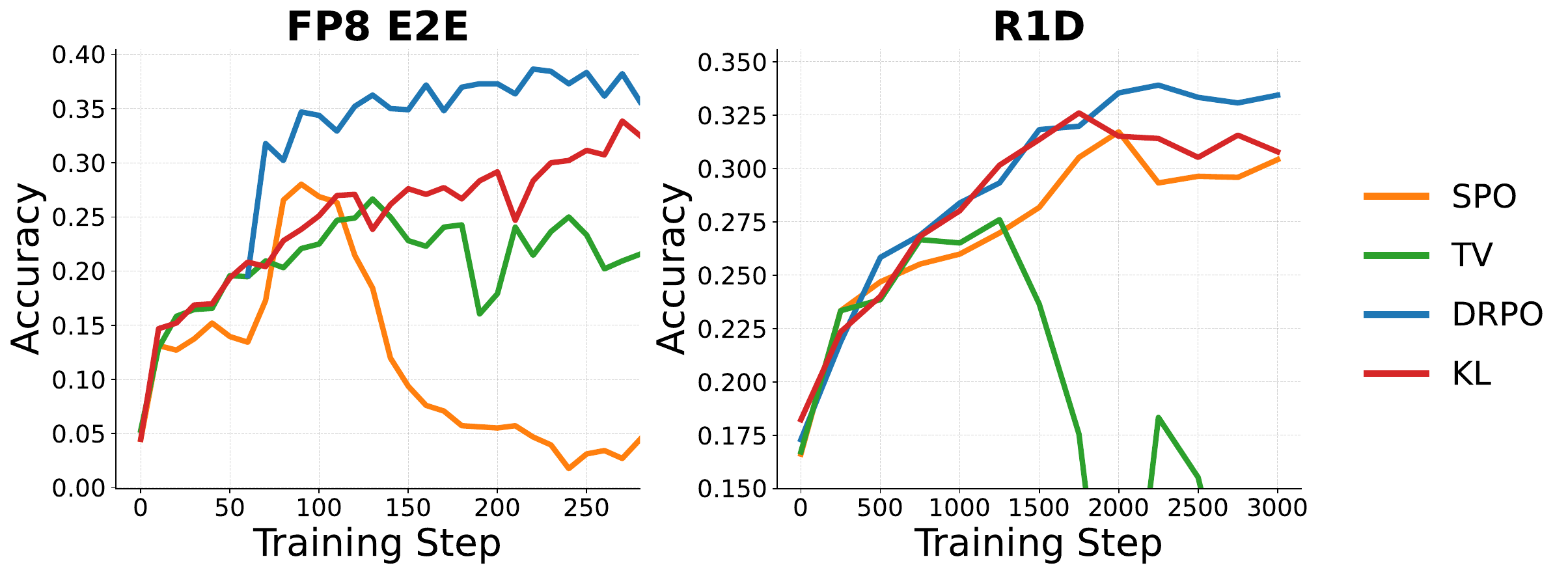}
    \caption{Ablation on alternative divergence metrics. \method{} achieves the best performance.}
    \label{fig:ablation_divergence}
\end{figure}

To answer this question, we compare \method{} with several alternatives, including commonly used forward KL and TV penalties (\Cref{eq:kl_penalty} and \ref{eq:tv_penalty}). As shown in \Cref{fig:ablation_divergence}, all of these alternatives underperform \method{}. We argue that this result is expected because their per-token gradients induce either binary or ratio-based optima rather than a smooth Binary-TV boundary, with the detailed analysis deferred to Appendix~\ref{sec:app:alternative_divergences}.
In contrast, \method{} induces a Binary-TV trust region, which provides more stable gradients and better captures the true distributional shift, as detailed in \Cref{sec:divergence_comparison}.

\noindent\textbf{Applying the regularizer only outside DPPO's trust region.}
To examine where the performance gain of \method{} primarily comes from, we conduct an experiment in which the regularizer is applied only outside the DPPO trust region. We refer to this variant as Mask-\method{}. Within the DPPO trust-region boundary, Mask-\method{} has the same gradient as DPPO; outside this boundary, it has the same gradient as \method{}. As shown in \Cref{fig:ablation_mask_fp8e2e_drpo_spo}, Mask-\method{} achieves performance comparable to \method{}, suggesting that the main performance gain comes from the corrective regularization outside the trust region. 
In addition, other regularizer alternatives still do not match \method{}'s performance, further supporting the effectiveness of our design. See Appendix~\ref{sec:app:mask_divergence_ablation} for more details.

\section{Closing Remarks}

Many prior works design regularizers from the objective perspective, typically by adopting standard divergence measures such as KL, JS, or related variants. Our empirical results suggest that the induced gradient form is more critical than the nominal divergence in the objective. For example, although the absolute-advantage term in \Cref{eq:spo_obj} and \Cref{eq:our_objective_mc} prevents the regularizer from being a pure divergence, we find that it is essential for maintaining a stable trust-region boundary and enabling stable training.

The choice of regularizer therefore requires careful consideration. A regularization term that appears reasonable at the objective level can perform poorly if its gradient induces undesirable geometry. In particular, we identify a common failure mode in which the gradient induces a ratio-based trust region, whose weights can have high variance and become unbounded under the long-tailed vocabularies of LLMs. In contrast, the absolute probability shift, namely Binary-TV, provides a better alternative: it is bounded and better captures the geometry of TV divergence.

This observation is consistent with DPPO~\citep{dppo}, which replaces ratio-based clipping in PPO with a divergence-based mask. However, DPPO still relies on a hard mask, whose effective gradient changes abruptly near the mask boundary and provides no corrective signal outside the trust region. To address this, we propose \method{}, which replaces the hard mask with a smooth quadratic regularizer while preserving the same trust-region geometry. Across dense and MoE architectures, reasoning and non-reasoning models, and BF16 and FP8 precision settings, \method{} improves training stability and achieves stronger performance than a diverse set of baselines.

\bibliography{ref}

\clearpage
\appendix
\section{Related Work}
\label{sec:related_work}

\subsection{Traditional RL based on Trust Region Methods}
Trust region methods ensure stable policy optimization by limiting how much the policy can change in each update. TRPO~\citep{trpo} derives a policy improvement bound penalized by TV divergence and solves the resulting KL-constrained optimization via conjugate gradient, guaranteeing monotonic improvement. CPO~\citep{cpo} extends this to constrained MDPs. However, both require second-order optimization that is prohibitive at scale.

PPO~\citep{ppo} replaces the explicit KL constraint with a ratio-clipping heuristic, enabling first-order optimization. Despite its success, the clipping mechanism neither strictly bounds the likelihood ratio nor enforces a well-defined divergence constraint~\citep{wang2020truly}. Truly PPO~\citep{wang2020truly} addresses this by introducing a rollback clipping function with a KL-based triggering condition. Trust Region-Guided PPO~\citep{wang2019trust} proposes adaptive clipping thresholds guided by KL divergence, providing stronger guarantees than fixed-width clipping.
MDPO~\citep{tomar2022mirror} connects trust-region policy optimization with mirror descent~\citep{beck2003mirror}, approximately solving the trust-region subproblem via multiple gradient steps on a Bregman divergence objective rather than enforcing a hard constraint.

Most relevant to our work, SPO~\citep{spo} replaces PPO's hard clipping with a smooth quadratic regularizer on the importance ratio. The per-token optimum of the resulting concave quadratic exactly matches PPO's clipping boundary, while providing non-zero corrective gradients outside the trust region. Our method adopts SPO's smooth regularization principle but changes the trust-region geometry from ratio-based to divergence-based. Specifically, we weight SPO's quadratic penalty by the behavior probability $\mu(y_t|s_t)$, which transforms the implicit regularization from a $\chi^2$-type penalty to an $\ell_2^2$-type penalty on probability shifts. This single modification changes the per-token optimum from the ratio boundary $|r_t - 1| = \epsilon$ to the Binary-TV boundary $|\pi(y_t|s_t) - \mu(y_t|s_t)| = \delta$, inheriting the smooth gradient structure of SPO while aligning the trust region with the TV geometry of DPPO.

\subsection{RL for LLM Reasoning}
Reinforcement learning has become a key technique for improving reasoning in LLMs~\citep{guo2025deepseekr1,team2025kimi}. In practice, LLM RL is inherently off-policy due to training-inference mismatch~\citep{yao2025offpolicy,fp16} and mini-batch policy staleness~\citep{deepseek-v3.2}, making trust-region optimization essential for stable training.

The dominant approach uses PPO-style hard clipping to impose ratio-based trust regions. GRPO~\citep{grpo,drgrpo} retains this objective while replacing critic-based advantages with group-relative advantages~\citep{drgrpo,zeng2025simplerl}. DAPO~\citep{yu2025dapo} asymmetrically widens the upper clipping bound, CISPO~\citep{cispo} removes clipping through truncated importance sampling, and M2PO~\citep{zheng2025prosperity} constrains the second moment of importance weights. To reduce variance under off-policy data, prior work has also proposed truncated~\citep{yao2025offpolicy,zheng2025stabilizing} and masked~\citep{liu-li-2025,team2025every} importance sampling.

Another line of work uses regularization to enforce trust-region behavior instead of relying on hard clipping or masking. Kimi k1.5~\citep{team2025kimi} and Kimi k2.5~\citep{team2026kimi} adopt online policy mirror descent. R$^2$VPO~\citep{r2vpo} replaces hard clipping with a smooth Lagrangian penalty on ratio variance, but it remains ratio-based and can induce unbounded gradient weights for low-probability tokens. TROLL~\citep{becker2025troll} enforces per-token KL constraints through differentiable projections, but requires solving an optimization problem for each token.

These mask-based and regularizer-based methods either remain tied to the importance ratio or adjust the trust region heuristically, without directly resolving the mismatch between ratio change and distributional shift. DPPO~\citep{dppo} identifies this flaw in long-tailed vocabularies~\citep{wang2025beyond} and replaces ratio clipping with a divergence-based binary mask on TV or KL divergence. However, DPPO still changes gradients abruptly at the boundary and provides no corrective signal once a token moves outside the trust region.

\method{} combines the divergence-based geometry of DPPO with the smooth enforcement principle used by regularizer-based methods such as R$^2$VPO, while avoiding their main limitations. \method{} preserves the directional structure of PPO and DPPO: it attenuates updates that move the policy away from the behavior policy and amplifies updates that move it back. Through a lightweight advantage-weighted $\ell_2^2$ regularizer, \method{} aligns the update with Binary-TV geometry, provides smooth corrective gradients, and keeps per-token gradient weights bounded.

\section{Detailed Derivation of the Gradient of \method{}}\label{app:full_grad_derivation}

The gradient of the objective in \Cref{eq:our_objective_mc} can be derived as follows:
\begin{equation*}\label{eq:full_grad}
    \begin{split}
        &\nabla \mathcal{L}_\text{\method}(x,\pi) \\
        ={}& \mathbb{E}_{y \sim \mu(\cdot|x)}\!\left[\sum_{t=1}^{|y|}
            \nabla r_t \hat{A}_t
            - \frac{|\hat{A}_t|}{\delta}\,\mu(y_t|s_t)(r_t - 1)\nabla r_t
        \right] \\
        ={}& \mathbb{E}_{y \sim \mu(\cdot|x)}\!\left[\sum_{t=1}^{|y|}
            \left(1 - \operatorname{sign}(\hat{A}_t)\frac{\mu(y_t|s_t)(r_t - 1)}{\delta}\right)
            \nabla r_t \hat{A}_t
        \right] \\
        ={}& \mathbb{E}_{y \sim \mu(\cdot|x)}\!\left[\sum_{t=1}^{|y|}
            \left(1 - \operatorname{sign}(\hat{A}_t(r_t - 1))\frac{|\pi(y_t|s_t)-\mu(y_t|s_t)|}{\delta}\right)
            r_t \hat{A}_t \nabla \log \pi(y_t|s_t)
        \right] \\
        ={}& \mathbb{E}_{y \sim \mu(\cdot|x)}\!\left[\sum_{t=1}^{|y|}
            \left(1 - \operatorname{sign}(\hat{A}_t(r_t - 1))\frac{D_t^{\mathrm{Bin\text{-}TV}}}{\delta}\right)
            r_t \hat{A}_t \nabla \log \pi(y_t|s_t)
        \right].
    \end{split}
\end{equation*}

\section{Induced Trust Regions of Alternative Regularizers}
\label{sec:app:alternative_divergences}

We analyze the trust region induced by each alternative regularizer through its per-token gradient. Fix a state $s_t$ and a sampled token $y_t$. For compactness, denote
\[
    \mu_t \triangleq \mu(y_t|s_t), \qquad
    \pi_t \triangleq \pi(y_t|s_t), \qquad
    r_t \triangleq \frac{\pi_t}{\mu_t}.
\]
This appendix is intended to clarify a subtle point in regularizer design. Two objectives can look similar at the loss level but induce very different gradient geometries after importance sampling. For LLM RL, this distinction is important because the optimization update is driven by sampled tokens from a highly long-tailed vocabulary. A useful trust-region regularizer should therefore be judged not only by the name of the divergence it resembles, but also by the scalar weight it applies to the token-level policy gradient.

We consider the following alternative regularizers:
\begin{align}
    \mathcal{L}_\mathrm{KL}(x, \pi) &= \mathbb{E}_{y \sim \mu(\cdot|x)}\left[  \sum_{t=1}^{|y|}  r_t \cdot \hat{A}_t +\frac{|\hat A_t|}{2\delta}\cdot \log r_t\right], \label{eq:kl_penalty} \\
    \mathcal{L}_\mathrm{K3}(x, \pi) &= \mathbb{E}_{y \sim \mu(\cdot|x)}\left[  \sum_{t=1}^{|y|}  r_t \cdot \hat{A}_t - \frac{|\hat A_t|}{2\delta}\cdot (r_t-1-\log r_t)\right], \label{eq:k3_penalty} \\
    \mathcal{L}_\mathrm{TV}(x, \pi) &= \mathbb{E}_{y \sim \mu(\cdot|x)}\left[  \sum_{t=1}^{|y|}  r_t \cdot \hat{A}_t -\frac{|\hat A_t|}{2\delta}\cdot |r_t-1|\right], \label{eq:tv_penalty}
\end{align}
Since $\mu_t$ is fixed during the policy update, $\nabla r_t=r_t\nabla\log\pi_t$. We therefore write each gradient as the original policy-gradient term $r_t\hat A_t\nabla\log\pi_t$ multiplied by an induced weight. The zero of this weight gives the boundary at which the regularizer cancels the reward-improving gradient.
When this boundary is expressed as a fixed value of $r_t$, the regularizer inherits the same ratio-based geometry as PPO and SPO. When the boundary is expressed as a fixed value of $|\pi_t-\mu_t|$, it matches the Binary-TV geometry used by \method{} and DPPO.

\noindent\textbf{Advantage-weighted KL regularizer.}
Consider the per-token KL-regularized objective
\begin{equation}
    \ell_{\mathrm{KL}}(r_t)
    =
    r_t\hat A_t
    +
    \frac{|\hat A_t|}{2\delta} \log r_t.
    \label{eq:app_kl_penalty}
\end{equation}
This is the sampled contribution of the forward KL penalty $D_{\mathrm{KL}}(\mu\|\pi)$ under the behavior-policy expectation, up to the sign convention induced by maximizing the objective. Taking the gradient gives
\begin{align}
    \nabla \ell_{\mathrm{KL}}(r_t)
    &=
    \left(
        r_t\hat A_t
        +
        \frac{|\hat A_t|}{2\delta}
    \right)\nabla\log\pi_t \notag\\
    &=
    \left(
        1
        +
        \frac{\operatorname{sign}(\hat A_t)}{2\delta r_t}
    \right)
    r_t\hat A_t\nabla\log\pi_t .
    \label{eq:app_kl_grad}
\end{align}
Thus the KL-induced gradient weight is
\[
    w_{\mathrm{KL}}(r_t)
    =
    1
    +
    \frac{\operatorname{sign}(\hat A_t)}{2\delta r_t}.
\]
A key observation is that the gradient weight only depends on $r_t$, which leads to a ratio-based geometry. Setting $w_{\mathrm{KL}}(r_t)=0$ yields
\begin{equation}
    r_t^\star
    =
    -\frac{\operatorname{sign}(\hat A_t)}{2\delta}.
    \label{eq:app_kl_boundary}
\end{equation}
For $\hat A_t>0$, this equation has no feasible solution because $r_t>0$ and the gradient weight is always positive. For $\hat A_t<0$, the zero-gradient point is
\[
    r_t^\star = \frac{1}{2\delta},
    \qquad
    \pi_t^\star = \frac{\mu_t}{2\delta}.
\]
Therefore, whenever the KL penalty induces a finite cancellation boundary, that boundary is ratio-based. The stopping condition depends on $\pi_t/\mu_t$, not on the absolute probability shift.
This also explains why directly adding a KL penalty is not a drop-in replacement for \method{}. For positive-advantage tokens, the sampled forward-KL term does not create a finite rollback point in this one-sample gradient form; for negative-advantage tokens, the rollback point scales with $\mu_t$. Consequently, a rare token and a frequent token can receive very different absolute probability tolerances even when their semantic effect on the next-token distribution should be judged by probability mass rather than by relative ratio.

\noindent\textbf{Advantage-weighted KL regularizer with the K3 estimator.}
The previous objective uses the K1 estimator $-\log r_t$ for $D_{\mathrm{KL}}(\mu\|\pi)$, which can have high variance. A common lower-variance alternative is the K3 estimator
\[
    k_3(r_t)=r_t-1-\log r_t,
\]
which has the same expectation under $y_t\sim\mu(\cdot|s_t)$ because $\mathbb{E}_{y_t\sim\mu}[r_t-1]=0$. The corresponding per-token objective is
\begin{equation}
    \ell_{\mathrm{KL3}}(r_t)
    =
    r_t\hat A_t
    -
    \frac{|\hat A_t|}{2\delta}
    \bigl(r_t-1-\log r_t\bigr).
    \label{eq:app_kl3_penalty}
\end{equation}
Taking the gradient gives
\begin{align}
    \nabla \ell_{\mathrm{KL3}}(r_t)
    &=
    \left(
        r_t\hat A_t
        -
        \frac{|\hat A_t|}{2\delta}(r_t-1)
    \right)\nabla\log\pi_t \notag\\
    &=
    \left(
        1
        -
        \frac{\operatorname{sign}(\hat A_t)(r_t-1)}{2\delta r_t}
    \right)
    r_t\hat A_t\nabla\log\pi_t .
    \label{eq:app_kl3_grad}
\end{align}
Thus the K3-induced gradient weight is
\[
    w_{\mathrm{KL3}}(r_t)
    =
    1
    -
    \frac{\operatorname{sign}(\hat A_t)(r_t-1)}{2\delta r_t},
\]
which also gives a ratio-based geometry because it only depends on $r_t$.
Setting $w_{\mathrm{KL3}}(r_t)=0$ yields
\[
    r_t^\star
    =
    \begin{cases}
        \frac{1}{1-2\delta}, & \hat A_t>0 \text{ and } \delta<\frac{1}{2},\\[0.6em]
        \frac{1}{1+2\delta}, & \hat A_t<0.
    \end{cases}
\]
For $\hat A_t>0$ and $\delta\ge\frac{1}{2}$, the gradient weight remains positive for all feasible $r_t>0$, so no finite cancellation boundary exists. When a finite boundary does exist, it is again expressed as a fixed value of the importance ratio $r_t=\pi_t/\mu_t$. The K3 estimator reduces the variance of the KL estimate, but it still induces a ratio-based trust region.
In other words, K3 changes the estimator but not the relevant geometry. It can make the KL estimate numerically better behaved, yet the corrective force is still calibrated in ratio space. This is the key mismatch for long-tailed language-model distributions: a small absolute movement on a low-probability token can dominate the gradient through the ratio factor $r_t$ as $r_t$ grows large, while a much larger movement on a high-probability token may appear modest in ratio terms.

\noindent\textbf{Advantage-weighted TV regularizer.}
Now consider the per-token TV-regularized objective
\begin{equation}
    \ell_{\mathrm{TV}}(r_t)
    =
    r_t\hat A_t
    -
    \frac{|\hat A_t|}{2\delta}|r_t-1|.
    \label{eq:app_tv_penalty}
\end{equation}
For $r_t\ne 1$, its gradient is
\begin{align}
    \nabla \ell_{\mathrm{TV}}(r_t)
    &=
    \left(
        r_t\hat A_t
        -
        \frac{|\hat A_t|}{2\delta}r_t\operatorname{sign}(r_t-1)
    \right)\nabla\log\pi_t \notag\\
    &=
    \left(
        1
        -
        \frac{\operatorname{sign}(\hat A_t)\operatorname{sign}(r_t-1)}{2\delta}
    \right)
    r_t\hat A_t\nabla\log\pi_t .
    \label{eq:app_tv_grad}
\end{align}
Thus the TV-induced gradient weight is
\begin{equation}
    w_{\mathrm{TV}}(r_t)
    =
    1
    -
    \frac{\operatorname{sign}(\hat A_t)\operatorname{sign}(r_t-1)}{2\delta}.
    \label{eq:app_tv_weight}
\end{equation}
This weight takes only two values:
\[
    w_{\mathrm{TV}}(r_t)
    =
    \begin{cases}
        1-\frac{1}{2\delta},
        & \operatorname{sign}\!\bigl(\hat A_t(r_t-1)\bigr)>0,\\[0.3em]
        1+\frac{1}{2\delta},
        & \operatorname{sign}\!\bigl(\hat A_t(r_t-1)\bigr)<0.
    \end{cases}
\]
It depends only on whether the current ratio shift has the same sign as the advantage. It does not depend on the magnitude of $|r_t-1|$.
The advantage-weighted TV penalty therefore induces a binary gradient weight, not a smooth trust-region boundary.

This behavior is undesirable for a different reason from KL. The TV penalty removes the unbounded ratio magnitude, but the sampled absolute-value form has a nondifferentiable kink at $r_t=1$ and a piecewise-constant gradient weight away from that point. As a result, it distinguishes only whether the update is moving away from or toward the behavior policy, not how far the token has moved. It therefore cannot reproduce the gradual attenuation inside the trust region or the strength-calibrated correction outside the boundary that \method{} provides.

\paragraph{Summary.}
The above derivations show that the nominal divergence in the objective is not sufficient to determine whether a method has the desired trust-region behavior. KL and K3 penalties induce ratio-based boundaries; the sampled TV penalty induces a two-level gradient weight; and none of them yields a smooth Binary-TV boundary. By contrast, the \method{} regularizer in \Cref{eq:our_objective_mc} produces the weight
\[
    1-\operatorname{sign}(\hat A_t(r_t-1))\frac{|\pi_t-\mu_t|}{\delta},
\]
which depends continuously on the absolute probability shift. This is the property that lets \method{} preserve DPPO's divergence-based trust-region geometry while replacing the hard mask with a corrective smooth update. The empirical comparisons in Appendix~\ref{sec:app:experiment} and Appendix~\ref{sec:app:mask_divergence_ablation} are consistent with this analysis: penalties whose gradients remain ratio-based or binary are less stable than the Binary-TV quadratic penalty.

\section{More Experimental Details}
\label{sec:app:experiment}

\begin{table}[h]
    \centering
    \caption{Hyperparameters.}
    \label{tab:hyperparameters}
    \resizebox{\textwidth}{!}{
    \begin{tabular}{lcccc}
    \toprule
        \textbf{Hyperparameters} & \textbf{Qwen3-4B-Base} & \textbf{Qwen3-30B-A3B-Base} & \textbf{Qwen3.5-35B-A3B-Base} & \textbf{R1D} \\\midrule
        Learning Rate & 1e-6 & 1e-6 & 1e-6 & 1e-6 \\
        PPO Epochs & 1 & 1 & 1 & 1 \\
        Max Prompt Length & 2048 & 2048 & 2048 & 2048 \\
        Max Response Length & 8192 & 8192 & 8192 & 8192 \\
        Train Batch Size & 64 & 256 & 256 & 64 \\
        PPO Mini Batch Size & 32 & 32 & 32 & 16 \\
        Rollout Temperature & 1.0 & 1.0 & 1.0 & 1.0 \\
        Group Size & 8 & 16 & 16 & 8 \\
    \bottomrule
    \end{tabular}}
\end{table}

We provide the detailed experiment configurations, more ablation studies, and results in this section as a complementary part of Section~\ref{sec:experiment}. 

For the computation resources, we use 4 $\times$ 8 NVIDIA H20 to conduct most of the experiments. We build our codebase on VeRL~\citep{sheng2024hybridflow} and use Megatron~\citep{megatron-lm} as the training backend and vLLM~\citep{kwon2023efficient} as the inference backend to speed up rollout. To verify the correctness of the solutions in math reasoning tasks, we utilize the third-party library math-verify\footnote{\url{https://github.com/huggingface/Math-Verify}}.

Besides, we have tried various kinds of objective functions, revealing the effects of the advantage scaling $|\hat A_t|$, different divergences, binary approximation, etc.  
Typically, we train Qwen3-4B-Base with 800 steps, Qwen3-30B-A3B-Base with 300 steps, Qwen3.5-35B-A3B-Base with 110 steps, and R1D with 3000 steps. Since we used Megatron \citep{megatron-lm} as the training backend, and at the time we conducted experiments, it did not have sufficient support for efficiently training Qwen3.5, we chose to train fewer steps compared to Qwen3-30B-A3B-Base.

\subsection{Comparing with KL Regularization}\label{app:comparing_kl}

\begin{figure}[h]
    \centering
    \includegraphics[width=0.9\linewidth]{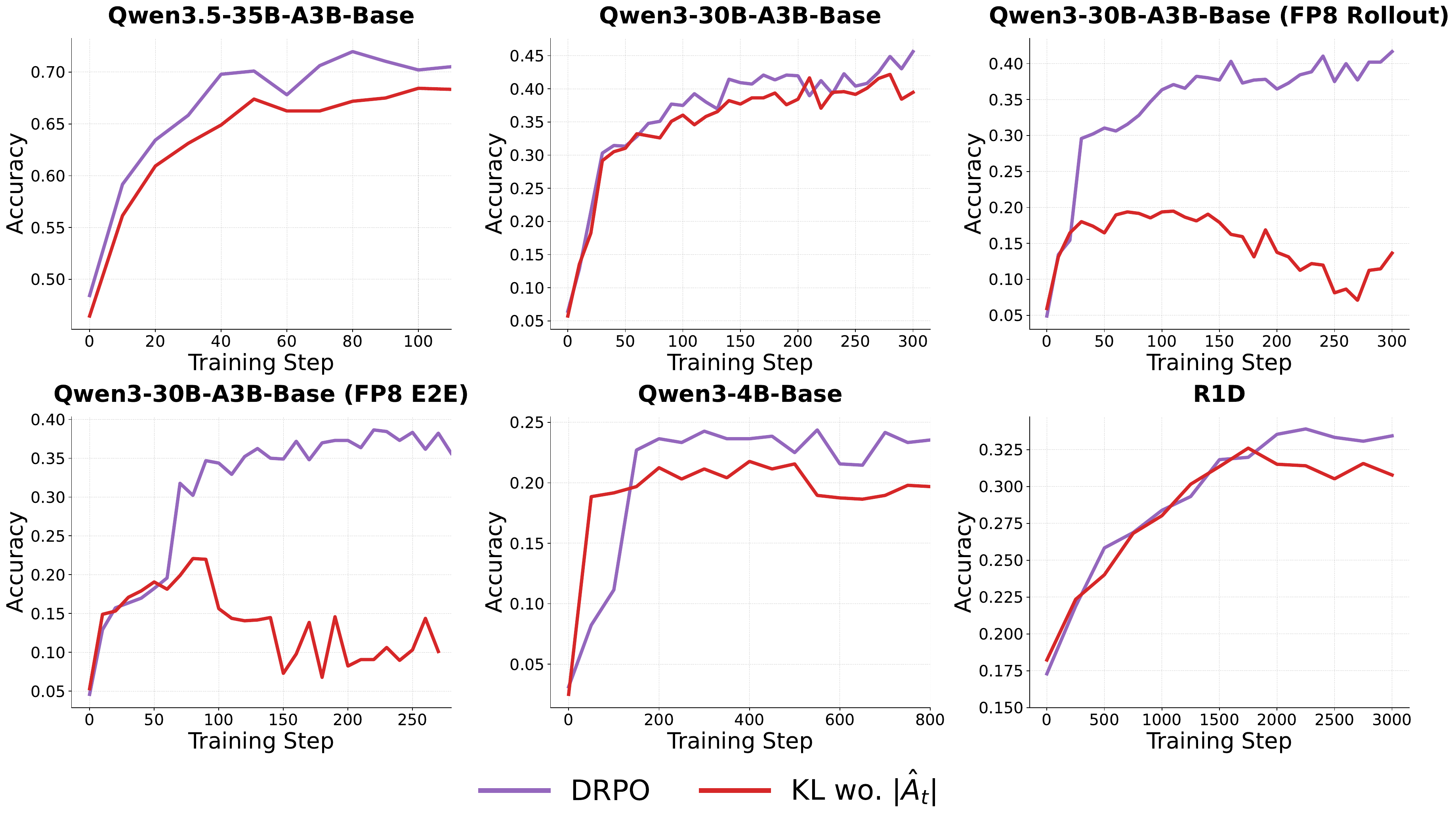}
    \caption{Training dynamics for \method{} and directly applying a KL penalty term without introducing the advantage weight $|\hat A_t|$.}
    \label{fig:drpo_vs_pgkl}
\end{figure}

In addition to the baselines in \Cref{sec:main_results}, another common method is to use a pure KL regularizer (without the advantage weight) as in the Algorithm 1 of \citet{trpo} and the Equation (8) of \citet{ppo}. We conduct an experiment to compare with this method. Specifically, we instantiate the below objective
\[
    \mathcal{L}_\mathrm{KL\_wo\_A}(x, \pi) = \mathbb{E}_{y \sim \mu(\cdot|x)}\left[  \sum_{t=1}^{|y|}  r_t \cdot \hat{A}_t +\frac{1}{2\delta}\cdot \log r_t\right]
\]
with the same hyperparameter $\delta=12.5$ (see the hyperparameter tuning results in \Cref{fig:r1d_kl_hparam}).

As shown in \Cref{fig:drpo_vs_pgkl}, \method{} consistently outperforms this KL regularizer across all six experiments. This gap can be explained from two complementary perspectives. First, the regularizer should adapt to the per-token advantage scale: as shown in Appendix~\ref{app:extended_adv_weight}, removing the factor $|\hat A_t|$ degrades performance because the token-wise optimum depends on the current advantage magnitude. Second, even after setting the advantage-weight issue aside, the KL penalty still induces a ratio-based trust-region geometry as analyzed in Appendix~\ref{sec:app:alternative_divergences}, which is less aligned with the desired constraint than \method{}.

\subsection{Extended Ablations on Advantage Weighting}\label{app:extended_adv_weight}

\begin{figure}[h]
    \centering
    \includegraphics[width=0.96\linewidth]{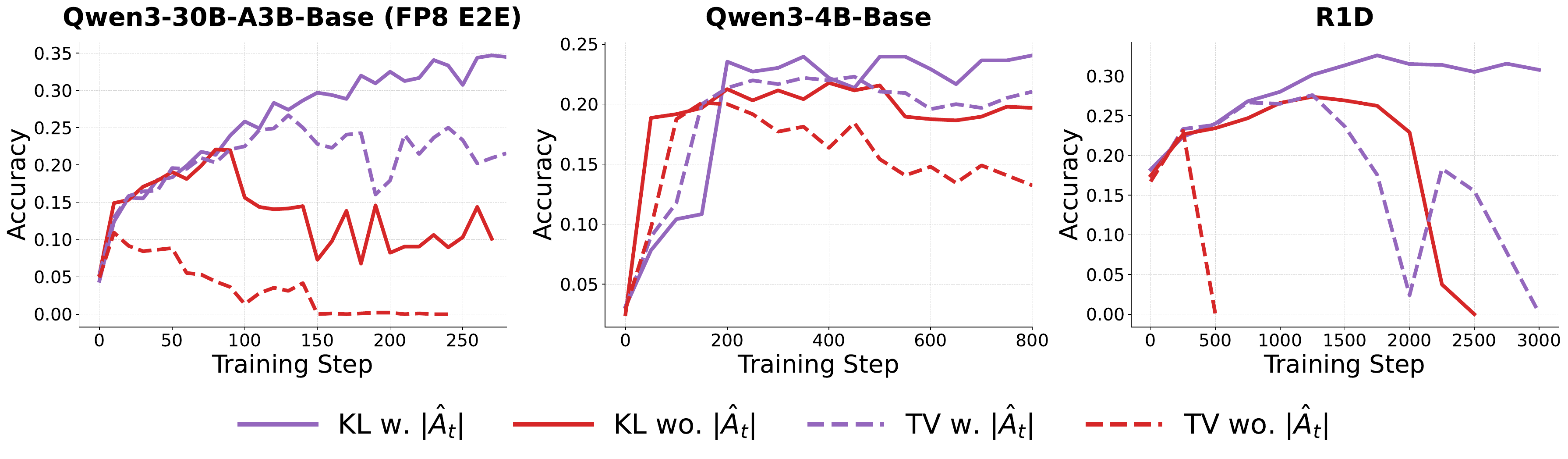}
    \caption{Comparison among experiments applying a KL penalty term or a TV penalty term, with or without the advantage weight $|\hat A_t|$.}
    \label{fig:wA_vs_woA}
\end{figure}

To further isolate the role of the advantage weight $|\hat A_t|$, \Cref{fig:wA_vs_woA} compares two penalty types, KL and TV, each evaluated both with and without this factor. The pattern is consistent across all settings: adding the advantage weight leads to clearly better performance, supporting the importance of weighting the regularizer by $|\hat A_t|$.

This behavior matches the analysis in \Cref{sec:trust_region_analysis} and the ablations in \Cref{sec:ablation_study}. Without $|\hat A_t|$, the effective trust-region boundary becomes entangled with the advantage magnitude rather than remaining stable. As a result, the update is overly restrictive for small-advantage tokens and too loose for large-advantage ones. Since token-level advantage estimates are also noisy in practice, this mismatch further hurts both training stability and final accuracy.

\subsection{Hyperparameter Tuning of Advantage-Weighted KL Regularizer}

\begin{figure}[h]
    \centering
    \includegraphics[width=0.9\linewidth]{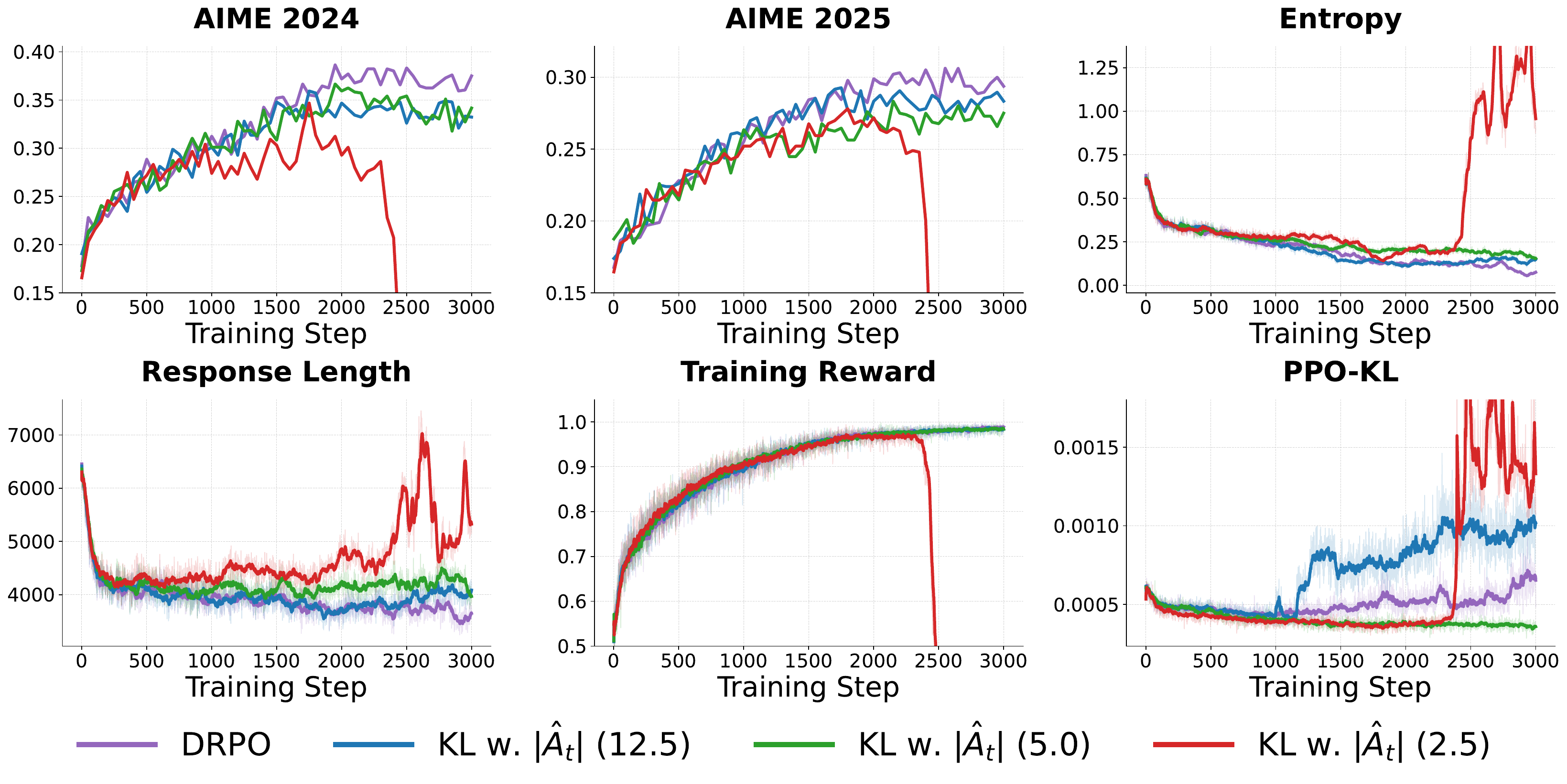}
    \caption{The hyperparameter tuning for KL with advantage weight $|\hat A_t|$ under the R1D setting.}
    \label{fig:r1d_kl_hparam}
\end{figure}

To rule out the concern that the KL baseline may simply be under-tuned, we sweep a range of hyperparameters for the advantage-weighted KL regularizer (\Cref{eq:k3_penalty}) under the R1D setting. Figure~\ref{fig:r1d_kl_hparam} shows that \method{} remains stronger across the full sweep, even when the KL baseline is equipped with the same advantage weight $|\hat A_t|$.

This robustness gap is consistent with our theoretical analysis. As discussed in Appendix~\ref{sec:app:alternative_divergences}, the KL penalty fundamentally imposes a ratio-based trust-region geometry. By contrast, \method{} yields a Binary-TV geometry, which more faithfully reflects the intended divergence constraint and therefore produces more reliable optimization behavior (see \Cref{sec:trust_region_analysis}).

\subsection{Hyperparameter Tuning of DPPO Baseline}

\begin{figure}[h]
    \centering
    \includegraphics[width=\linewidth]{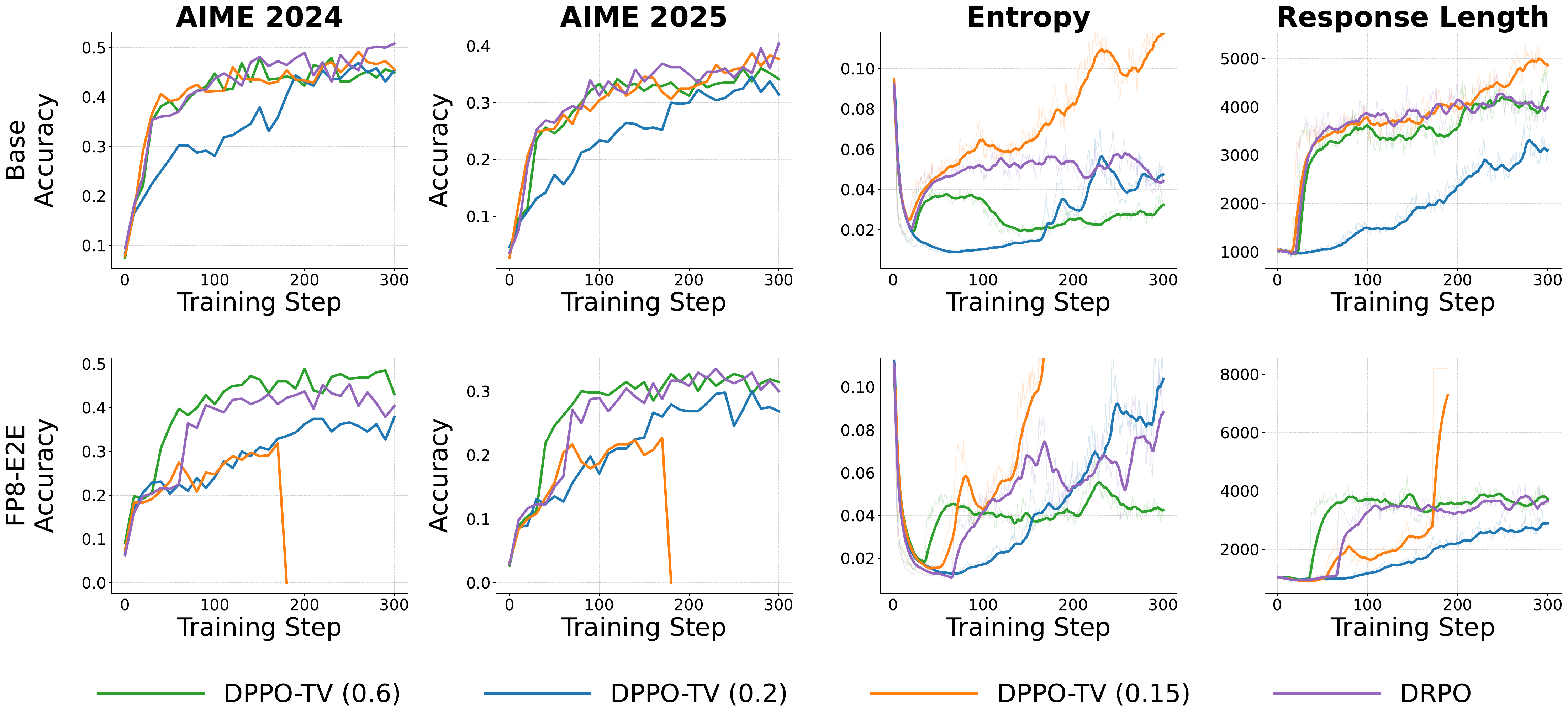}
    \caption{Training dynamics for different parameters under DPPO, compared to \method{}. \textbf{Top:} Qwen3-30B-A3B-Base; \textbf{Bottom:} Qwen3-30B-A3B-Base using FP8 precision for end-to-end training.}
    \label{fig:dppo_tv_tuning_param}
\end{figure}

To compare \method{} against a carefully tuned DPPO baseline, we sweep several DPPO thresholds. Unlike DPPO, \method{} provides corrective gradients for tokens outside the trust region.
Figure~\ref{fig:dppo_tv_tuning_param} shows that DPPO needs a more fine-grained parameter tuning, and $\varepsilon=0.15$ works best on the Qwen3-30B-A3B-Base setting, which still performs worse than \method{}, while $\varepsilon=0.6$ works best on the FP8-E2E setting, which achieves similar performance with \method{}. So \method{} has a relatively universal hyperparameter $\delta=12.5$ compared to DPPO.

\begin{figure}[h]
    \centering
    \includegraphics[width=0.9\linewidth]{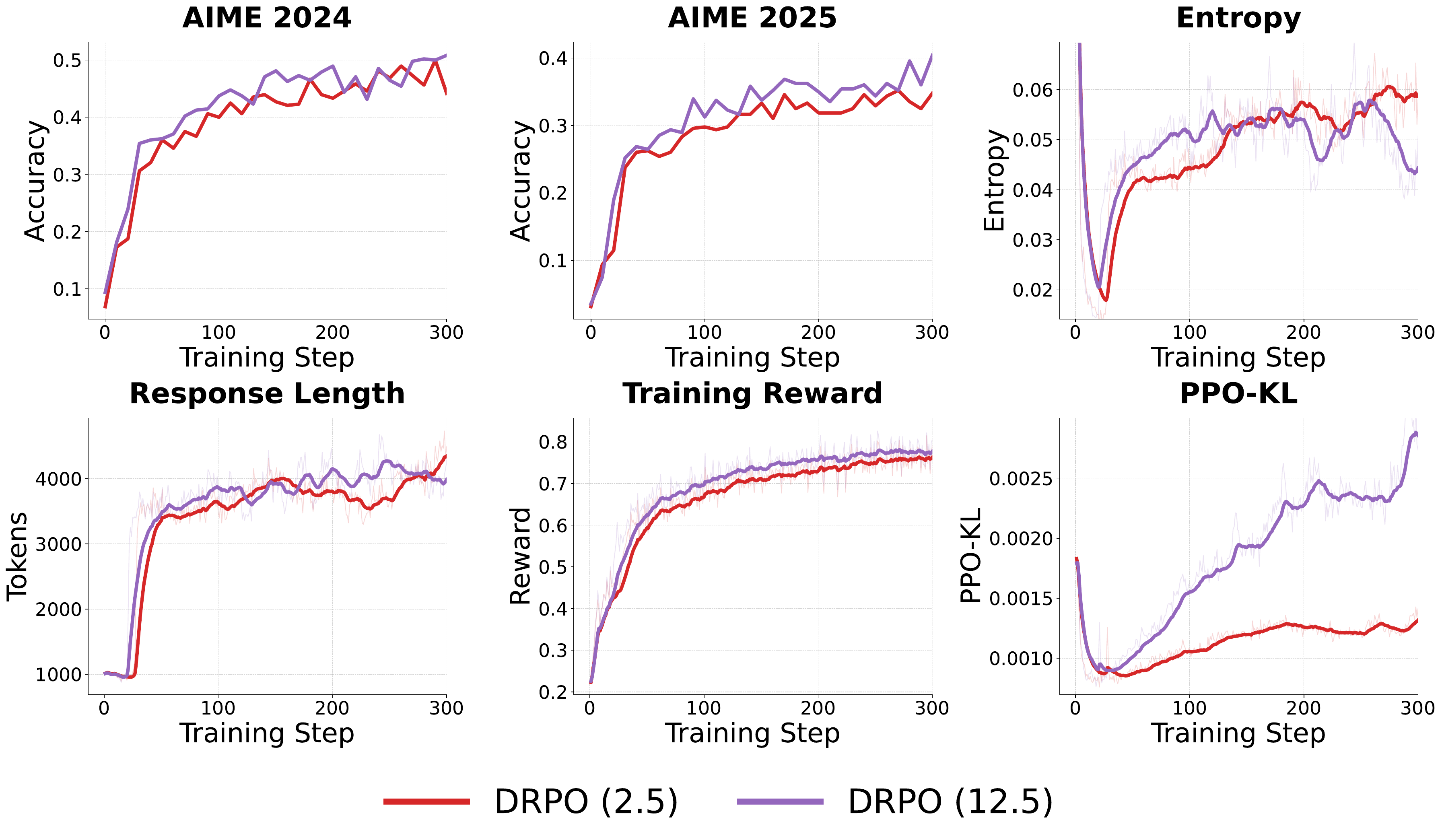}
    \caption{Hyperparameter tuning of the coefficient on \method{}.}
    \label{fig:drpo_tuning_param}
\end{figure}

\subsection{Hyperparameter Tuning of \method{}}


To assess the hyperparameter sensitivity of \method{}, we evaluate two choices of the threshold parameter $\delta$. As shown in \Cref{fig:drpo_tuning_param}, reducing $\delta$ substantially from 12.5 to 2.5 leads to only a minor drop in performance. This result suggests that \method{} is relatively robust to the choice of threshold and performs well across a broad hyperparameter range.

\subsection{Mask Ablation with Alternative Divergence Penalties}
\label{sec:app:mask_divergence_ablation}

We further repeat the mask ablation with several choices of divergence penalty. Let
\(
M_t^{\mathrm{out}}=\Id[D_t^{\mathrm{Bin\text{-}TV}}>\delta]
\)
denote the indicator that the sampled token is outside the DPPO trust region. For a generic penalty $\Omega_t$, the masked variant uses
\begin{equation*}
    \mathcal{L}_{\mathrm{Mask}\text{-}\Omega}(x,\pi)
    =
    \mathbb{E}_{y\sim\mu(\cdot|x)}
    \left[
        \sum_{t=1}^{|y|}
        r_t\hat A_t
        -
        M_t^{\mathrm{out}}\, \frac{|\hat A_t|}{2\delta}\,\Omega_t
    \right].
\end{equation*}
Inside the trust region, this objective reduces to the unregularized token surrogate, matching DPPO's active gradient. Outside the trust region, it restores the corrective penalty gradient that DPPO's hard mask discards.
Under this framework, we instantiate
three objectives as follows
\begin{align}
\label{eq:mask_drpo_objective}
    \mathcal{L}_{\mathrm{Mask\text{-}\method}}(x,\pi)
    &=
    \mathbb{E}_{y\sim\mu(\cdot|x)}
    \left[
        \sum_{t=1}^{|y|}
        r_t\hat A_t
        -
        M_t^{\mathrm{out}}\,
        \frac{|\hat A_t|}{2\delta}\,
        \mu(y_t|s_t)(r_t-1)^2
    \right],\\
\label{eq:mask_spo_objective}
    \mathcal{L}_{\mathrm{Mask\text{-}SPO}}(x,\pi)
    &=
    \mathbb{E}_{y\sim\mu(\cdot|x)}
    \left[
        \sum_{t=1}^{|y|}
        r_t\hat A_t
        -
        M_t^{\mathrm{out}}\,
        \frac{|\hat A_t|}{2\delta}\,
        (r_t-1)^2
    \right],\\
\label{eq:mask_kl_objective}
    \mathcal{L}_{\mathrm{Mask\text{-}KL}}(x,\pi)
    &=
    \mathbb{E}_{y\sim\mu(\cdot|x)}
    \left[
        \sum_{t=1}^{|y|}
        r_t\hat A_t
        -
        M_t^{\mathrm{out}}\,
        \frac{|\hat A_t|}{2\delta}\,
        \frac{(\log r_t)^2}{2}
    \right].
\end{align}
Notably, $(\log r_t)^2$ in \Cref{eq:mask_kl_objective} is exactly the penalty term in online policy mirror descent that Kimi series \citep{team2025kimi,team2026kimi} utilized.

\begin{figure}[h]
    \centering
    \includegraphics[width=0.6\linewidth]{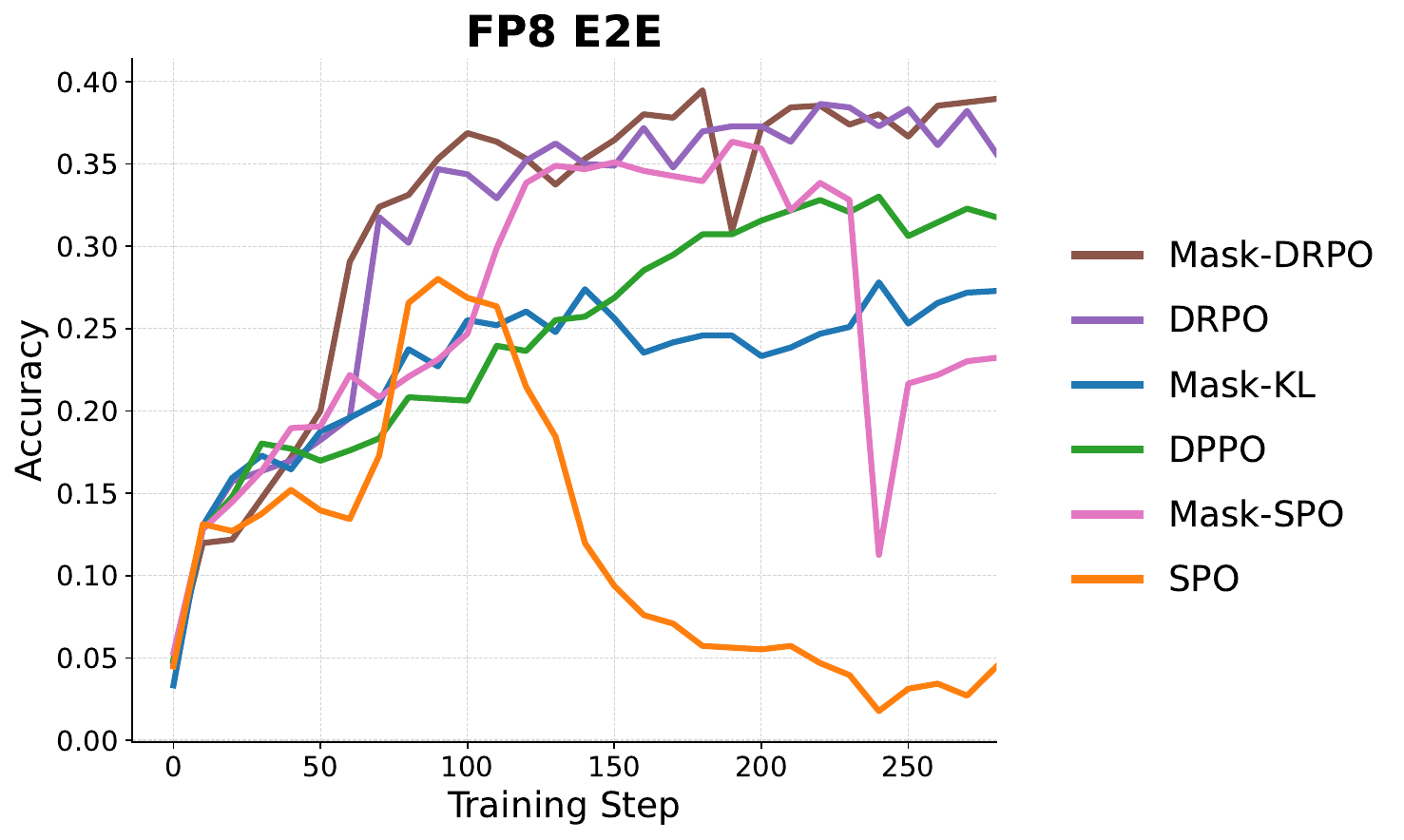}
    \caption{Ablation on applying the \method{} regularizer only outside DPPO's trust region.}
    \label{fig:ablation_mask_fp8e2e_drpo_spo}
\end{figure}

As shown in \Cref{fig:ablation_mask_fp8e2e_drpo_spo}, applying the penalty only outside the trust region achieves performance close to applying it everywhere for the \method{} regularizer, confirming that the main gain comes from correcting tokens that have crossed the boundary. At the same time, the penalty choice still matters: ratio-space and KL-type penalties are harder to calibrate, whereas the Binary-TV quadratic penalty used by \method{} gives the best and most stable behavior because its gradient directly follows absolute probability displacement.






\end{document}